\documentclass[runningheads]{llncs}

\usepackage[FINAL,year=2026]{eccv}

\usepackage{adjustbox}
\usepackage{pifont}

\usepackage{eccvabbrv}
\usepackage{makecell} 
\usepackage{graphicx}
\usepackage{booktabs}
\usepackage{multirow}
\usepackage[table]{xcolor}
\definecolor{tableblue}{RGB}{235, 245, 255}
\usepackage{graphicx}
\usepackage[accsupp]{axessibility}  
\usepackage{algorithm}
\usepackage{algorithmic}

\usepackage{hyperref}

\usepackage{orcidlink}

\usepackage{tcolorbox}
\usepackage{listings}
\usepackage{xcolor}
\usepackage{courier} 
\definecolor{varcolor}{rgb}{0.8, 0.2, 0.2}
\definecolor{promptbg}{rgb}{0.95, 0.95, 0.95}

\lstdefinestyle{promptstyle}{
    backgroundcolor=\color{promptbg},
    basicstyle=\ttfamily\scriptsize,
    breaklines=true,
    frame=none,
    showspaces=false,
    showstringspaces=false,
    showtabs=false,
    keywordstyle=\color{blue},
    stringstyle=\color{varcolor},
    commentstyle=\color{gray},
}

\newcommand{\promptvar}[1]{\textcolor{varcolor}{\texttt{\{#1\}}}}

\usepackage{tcolorbox}
\usepackage{listings}
\usepackage{xcolor}
\usepackage{courier} 
\definecolor{varcolor}{RGB}{122, 187, 219}
\definecolor{promptbg}{rgb}{0.95, 0.95, 0.95}

\lstdefinestyle{promptstyle}{
    backgroundcolor=\color{promptbg},
    basicstyle=\ttfamily\scriptsize,
    breaklines=true,
    frame=none,
    showspaces=false,
    showstringspaces=false,
    showtabs=false,
    keywordstyle=\color{blue},
    stringstyle=\color{varcolor},
    commentstyle=\color{gray},
}

\newtcolorbox{promptbox}[3][Judge Prompt]{
colback=black!5!white,
arc=5pt, 
boxrule=0.5pt,
fonttitle=\bfseries,
title=#1, 
before upper={\small}, fontupper=\fontfamily{ptm}\selectfont,
colframe=#2,
label=#3,
}
\definecolor{lightorange}{RGB}{253, 208, 162} 

\newtcolorbox{promptbox*}[3][Judge Prompt]{
  float*=ht,              
  width=\textwidth,       
  colback=black!5,
  colframe=#2,
  arc=5pt,
  boxrule=0.5pt,
  fonttitle=\bfseries,
  title=#1,
  before upper={\small},
  fontupper=\fontfamily{ptm}\selectfont,
  label=#3,
}

\begin{document}

\title{Reflect to Inform: Boosting Multimodal Reasoning via Information-Gain-Driven Verification} 

\titlerunning{Boosting Multimodal Reasoning via Information-Gain-Driven Verification}

\author{Shuai Lv\inst{1\ast}
\and
Chang Liu\inst{1\ast} \and
Feng Tang\inst{2\dag\#} \and
Yujie Yuan\inst{2} \and
Aojun Zhou\inst{3} \and
Kui Zhang\inst{2} \and
Xi Yang\inst{4\dag} \and
Yangqiu Song\inst{4} 
}

\authorrunning{S. Lv,~C. Liu et al.}

\institute{
University of Science and Technology of China, Hefei, China\\
\and
Huawei Foundation Model Department, Shanghai, China \\
\and
The Chinese University of Hong Kong, Hong Kong, China\\
\and
Hong Kong University of Science and Technology, Hong Kong, China
}

\maketitle

{\let\thefootnote\relax\footnotetext{
{$^{\ast}$ Equal contribution.~$\dag$ Corresponding authors.~$^{\#}$ Project Leader.}}}

\begin{abstract}
  Multimodal Large Language Models (MLLMs) achieve strong multimodal reasoning performance, yet we identify a recurring failure mode in long-form generation: as outputs grow longer, models progressively drift away from image evidence and fall back on textual priors, resulting in ungrounded reasoning and hallucinations.
  Interestingly, Based on attention analysis, we find that MLLMs have a latent capability for late-stage visual verification that is present but not consistently activated.
  Motivated by this observation, we propose Visual Re-Examination (VRE), a self-evolving training framework that enables MLLMs to autonomously perform visual introspection during reasoning without additional visual inputs. 
 Rather than distilling visual capabilities from a stronger teacher, VRE promotes iterative self-improvement by leveraging the model itself to generate reflection traces, making visual information actionable through information gain.
  Extensive experiments across diverse multimodal benchmarks demonstrate that VRE consistently improves reasoning accuracy and perceptual reliability, while substantially reducing hallucinations, especially in long-chain settings. Code is available at \url{https://github.com/Xiaobu-USTC/VRE}.
\end{abstract}

\section{Introduction}

\begin{figure*}[t]
\centering
\includegraphics[width=1\linewidth]{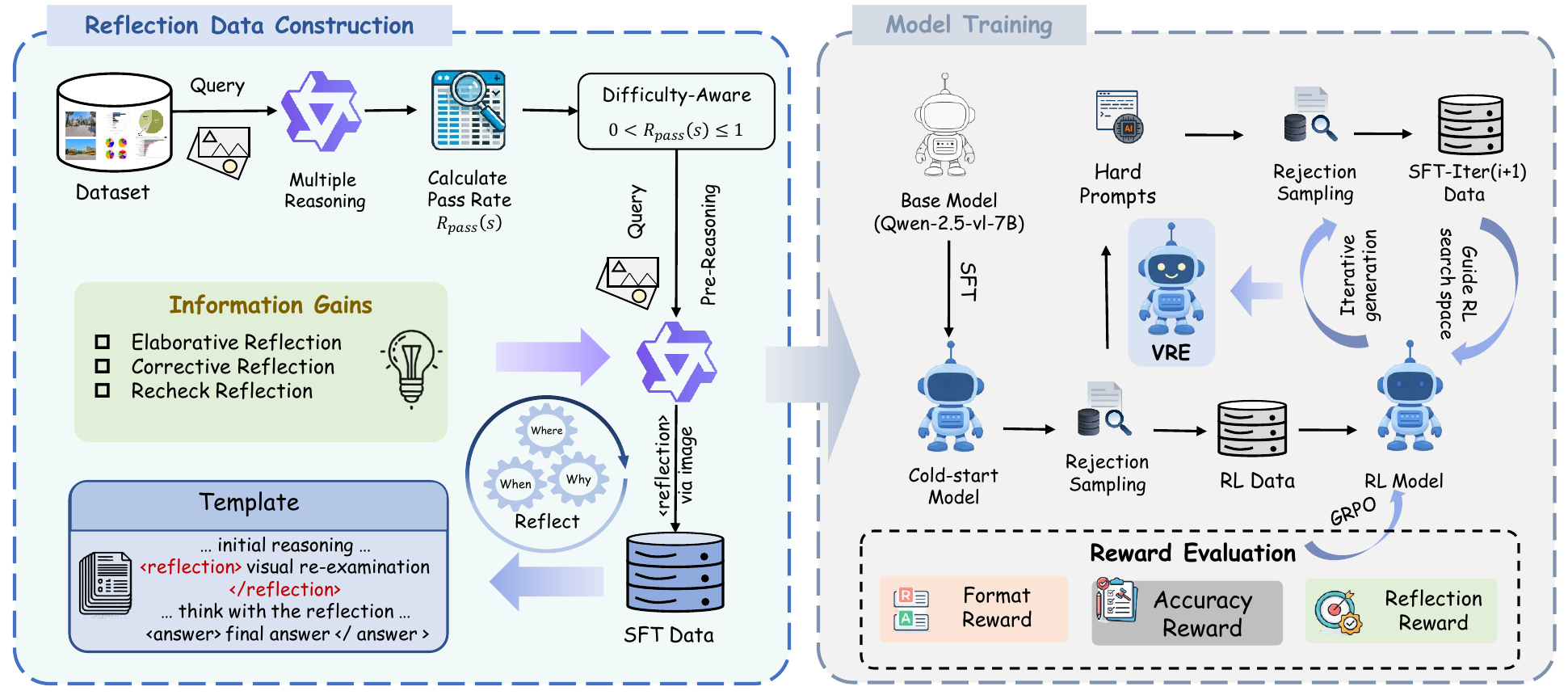}
\vspace{-15pt}

\caption{Overall pipeline of the proposed Visual Reflection Enhancement (VRE) framework. 
Left: Reflection data construction, where difficulty-aware filtering and information-gain-guided reflection synthesis generate high-quality SFT data. 
Right: Model optimization, where the cold-start model is 
iteratively improved through rejection sampling and 
reinforcement learning under a structured reward scheme 
comprising format, accuracy, and reflection rewards.}
\label{fig:pipline}
\vspace{-15pt}
\end{figure*}

Driven by the synergistic advancements in Multimodal Large Language Models (MLLMs)~\cite{liu2023visual,qwen2.5-VL,Qwen2VL,liao2025langbridge,lin2025uniworld} and reinforcement learning with verifiable rewards (RLVR)~\cite{shao2024deepseekmath, guo2025deepseek, meng2025mmeureka}, the boundary of vision-language problem-solving has been significantly expanded~\cite{amizadeh2020neuro, garcez2019neural, gupta2023visual, thawakar2025llamav, guo2024mammoth, bai2023qwen, hurst2024gpt, xu2024llava}. Despite these remarkable capabilities, we identify a persistent and critical bottleneck in extended multimodal generation, which we term visual drift~\cite{su2025openthinkimg, zhang2025chain, yang2025r1, hu2024visual, liu2025visual}. This progressive modality disconnect becomes exceptionally detrimental in scenarios necessitating late-stage visual re-verification, such as extracting fine-grained text or deducing intricate spatial layouts, ultimately culminating in unsubstantiated claims and severe hallucinations~\cite{liu2025more}.

To counteract this progressive modality disconnect, conventional methodologies predominantly resort to explicit visual interventions—such as dynamically injecting high-resolution patches~\cite{zheng2025deepeyes, hong2025deepeyesv2, zhang2025thyme, chern2025thinking, su2025openthinkimg}, superimposing visual markers~\cite{yang2023set,cai2024vip}, or interleaving new visual tokens~\cite{chen2025mint}. While these methods mitigate visual drift, they inevitably disrupt the standard auto-regressive input pipeline and impose substantial computational overhead. Alternatively, recent research reveals that MLLMs possess a latent capability for implicit visual refocusing, allowing them to spontaneously re-attend to original visual tokens without external additions~\cite{yang2025look}. Nevertheless, a critical bottleneck persists across both explicit and naive implicit paradigms: they typically lack strict guidance on what specific visual evidence to retrieve. Without a principled supervision signal, the model's re-look behavior frequently degenerates into redundant textual self-paraphrasing rather than actively capturing missing visual clues. Therefore, the core challenge is not merely enabling the model to relook, but transforming this unguided introspection into a purpose-driven mechanism.

Beyond this overall decay, we identify a more fundamental instability: for the same question, an MLLM can be ``visually clear'' in some instances but ``visually blind'' in others~\cite{guan2024hallusionbench, tong2024eyes}. Through preliminary experiments, we discover that under appropriate prompt guidance, MLLMs can spontaneously re-allocate attention to original image tokens in later decoding stages without any explicit visual re-injection~\cite{wu2024accelerating, lin2025boosting}. We term this latent capability Implicit Visual Re-Examination. This observation suggests that the ability to re-focus on visual evidence is not absent, but rather dormant and unreliably triggered~\cite{wu2024controlmllm, zhang2023multimodal}.

How can we turn this occasional, prompt-dependent re-examination 
into a robust and learnable policy? 
A prevailing paradigm for eliciting vision reasoning capabilities 
is strong-to-weak distillation from proprietary 
models~\cite{deng2025openvlthinker,yang2025look, zhang2025thyme,yeo2025demystifying}. 
As an alternative, we explore a different path: enabling MLLMs 
to self-evolve—learning autonomously when to initiate an implicit 
look-back and how to extract missing visual evidence.

To this end, we propose Visual Re-Examination (VRE), 
a self-iterative framework that strengthens implicit visual 
re-attention during long-form reasoning. VRE requires no architectural changes, additional visual inputs, 
or external teacher models. Instead, the model improves 
through iterative self-training, transforming sporadic 
re-focusing into a stable and learnable policy.

VRE employs an evidence-driven iterative framework: starting from a base model, we generate self-reflective reasoning traces and curate them via Reflection Information Gain, using rejection sampling to retain only traces with actionable and corrective visual evidence. These high-quality samples establish VRE formats via SFT, followed by RLVR that rewards both final answers and intermediate reflection quality. Iterating this RLVR-SFT pipeline enables the model to self-directively re-attend to informative visual regions and correct its reasoning trajectory.

\textbf{A pivotal design choice is the homologous reconstruction of reasoning trajectories.} Unlike existing works, we extract visual reasoning directly from the model's own trajectory, synthesizing reflections by concatenating this trace with the original multimodal input and re-feeding it into the same policy model for a secondary inference pass. Critically, employing the same visual backbone ensures strict perception-reasoning alignment, mitigating the disconnection risks inherent in approaches that leverage external knowledge beyond the model's native visual encoder. This homologous design prevents perceptual discrepancies (analogous to interpreting stimuli through ``another observer's eyes'') and biases stemming from mismatched visual features, ensuring that self-generated reflections remain consistent with the model's inherent visual understanding. We validate this design through analysis in 
Section~\ref{sec:reflction_ana}.

In this work, our key contributions are summarized as follows: 
\begin{itemize}
    \item We identify the phenomenon of visual drift in long-form multimodal reasoning and reveal the latent capability of Implicit Visual Re-Examination, shifting the paradigm from costly explicit visual re-injection to eliciting native visual attention.
    
    \item We propose VRE, a novel framework driven by a Self-Iterative Training Strategy. By enforcing perception-reasoning alignment through homologous reconstruction, VRE avoids the perceptual mismatch caused by relying on external knowledge sources inaccessible to the model's own visual encoder, enabling autonomous and consistent self-evolution of visual reasoning capabilities.
    
    \item We introduce \textit{Reflection Information Gain} within an evidence-driven data engine, effectively filtering noisy traces through rejection sampling to ensure that self-generated reflections provide genuine visual grounding.
    
    \item Extensive experiments demonstrate that VRE delivers consistent, iterative gains on diverse multimodal benchmarks, significantly enhancing long-chain visual reasoning while mitigating hallucinations.
\end{itemize}

\section{Related Work}
\subsection{Reasoning with Visual Information}
Executing complex multimodal tasks requires MLLMs to maintain sustained attention on visual evidence throughout extended reasoning chains~\cite{zhang2025chain, wang2025visuothink}. However, recent studies reveal a prevalent ``visual drift'' phenomenon: as the generated text lengthens, models increasingly rely on their internal language priors, neglecting the original image and producing ungrounded hallucinations~\cite{amizadeh2020neuro, garcez2019neural, thawakar2025llamav, bai2023qwen,hurst2024gpt, xu2024llava}.
To mitigate this drift, previous methodologies have predominantly adopted \textit{explicit intervention} strategies. These include dynamically injecting high-resolution image patches~\cite{zheng2025deepeyes, hong2025deepeyesv2, zhang2025thyme}, superimposing visual prompts like bounding boxes~\cite{yang2023set,cai2024vip}, or directly interleaving token-level visual features into textual reasoning steps~\cite{chen2025mint}. While effective, explicitly injecting new visual signals inevitably alters the input processing pipeline, requires complex heuristic designs, and incurs substantial computational overhead. 
Conversely, recent research demonstrates that MLLMs possess a latent capability to implicitly refocus on visual information without requiring any additional visual inputs~\cite{yang2025look}. Nevertheless, a critical limitation persists across both explicit and naive implicit approaches: they typically lack strict guidance on what specific visual evidence to retrieve. Without such direction, the model's re-look behavior often degenerates into redundant self-paraphrasing rather than capturing missing visual clues.

\subsection{Post-Training for MLLMs}
Post-training optimization has emerged as the de facto standard for unlocking complex reasoning capabilities in large models. Spurred by the paradigm-shifting success of OpenAI's o1~\cite{jaech2024openai}, the field has increasingly transitioned from module-centric designs to data-driven approaches, heavily leveraging SFT on large-scale synthetic chain-of-thought (CoT) trajectories. Beyond SFT, Reinforcement Learning (RL) has transcended its initial applications in human preference alignment~\cite{rlhfanthropic, oairlhf}. Recent breakthroughs demonstrate that Reinforcement Learning with Verifiable Rewards (RLVR)~\cite{guo2025deepseek, GRPO, tulu3} can profoundly elevate reasoning proficiency in structured domains such as mathematics and code generation.
However, directly adapting RLVR to Multimodal Large Language Models (MLLMs) presents unique and critical bottlenecks. Extensive RL fine-tuning in vision-language tasks frequently induces severe performance degradation. This decay is primarily driven by over-optimization~\cite{gao2023scaling} and multimodal reward hacking~\cite{casper2023open}, where the model learns to game the verification system by generating verbose textual illusions while entirely ignoring the actual visual input. Furthermore, prolonged RL trajectories often exacerbate the catastrophic forgetting of foundational, generic visual capabilities~\cite{zhai2023investigating}.

Our proposed framework elegantly circumvents these post-training pitfalls through a Self-Iterative Training Strategy. To fundamentally prevent textual reward hacking and ensure that intermediate reasoning steps remain tightly anchored to the image, we introduce Reflection Information Gain as the central criterion for both data curation and reward formulation. By rigorously penalizing redundant self-paraphrasing and explicitly rewarding the extraction of novel visual evidence, VRE guarantees that the post-training optimization strictly reinforces authentic, image-grounded deduction. This carefully guided regularization enables the MLLM to safely and continuously self-evolve its reasoning boundaries without succumbing to visual degradation.

\section{Method}
\label{sec:method}

Motivated by recent findings that MLLMs possess a latent capability for implicit visual re-examination~\cite{yang2025look}, we propose Visual Re-Examination (VRE), an iterative self-evolution framework. As illustrated in Fig.~\ref{fig:pipline}, VRE trains the MLLM to autonomously trigger a \texttt{<reflection>} process during long-form reasoning, re-attending to visual tokens to extract missing evidence. 

Our pipeline circumvents the pitfalls of strong-to-weak distillation by operating as a closed-loop self-iterative paradigm. It consists of three primary phases: (1) Reflection Data Construction for Cold-Start SFT; (2) Reinforcement Learning with Verifiable Rewards; and (3) Post-RL Self-Distillation to close the loop.

\subsection{Reflection Data Construction For Cold-Start}
\label{sec:coldstart}

The objective of this phase is to elicit and standardize the policy model's latent visual reflection behavior using its own distribution. Given a base MLLM $\pi_\theta$, we build a high-quality reflection data construction pipeline for cold-start SFT as shown in Fig.~\ref{fig:pipline}.

\paragraph{Step 1: Difficulty-Aware Sample Partitioning.}
Given a multimodal sample \( s = (i, q) \), 
where \( i \) denotes the image, \( q \) the question, 
and \( a_s \) the ground-truth answer, 
we draw \( N \) independent rollout trajectories 
from the base policy \( \pi_\theta \) using stochastic 
autoregressive sampling with a non-zero temperature. 
Each rollout trajectory \( \tau \) is a full generated sequence 
including intermediate reasoning tokens and a final answer: $
\tau = (y_1, y_2, \dots, y_T), 
\quad \tau \sim \pi_\theta(\cdot \mid s).
$ For each sample \( s \), the rollout set is defined as    $\mathcal{T}(s)
=
\{ \tau_i \}_{i=1}^{N}, \tau_i \sim \pi_\theta(\cdot \mid s).$
Let $\tau_{\text{ans}}$ denote the extracted final answer of trajectory \( \tau \). We define the pass rate as 
\begin{align}
P_{\text{pass}}(s)
=
\frac{1}{N}
\sum_{\tau \in \mathcal{T}(s)}
\mathbf{1}\!\left[\tau_{\text{ans}} = a_s\right]
\end{align}
which estimates the probability that the model produces 
the correct answer under its own sampling distribution. 
This quantity serves as a proxy for the model’s intrinsic competence on sample \( s \). Based on \( P_{\text{pass}}(s) \), we partition the dataset into three regimes:

\textbf{Stable} (\( P_{\text{pass}}(s) = 1 \)).  
The model succeeds consistently across all rollouts. These samples are retained to establish format priors and preserve previously mastered visual perception and reasoning ability, thereby mitigating catastrophic forgetting.

\textbf{Intractable} (\( P_{\text{pass}}(s) = 0 \)).  
The model fails on all rollouts. This design choice fundamentally differentiates our self-evolution framework from standard strong-to-weak distillation~\cite{yang2025look,zhang2025thyme,hong2025deepeyesv2}. Instead of forcing the model to imitate stronger MLLM teachers (e.g., GPT-5 or Gemini 2.5 Pro) on problems that exceed its intrinsic visual reasoning capacity, which may introduce the teacher’s visual inductive bias and distribution shift, we instead focus on unlocking the model’s own latent potential.

\textbf{Unstable} (\( 0 < P_{\text{pass}}(s) < 1 \)). The model exhibits inconsistent performance across rollouts. These queries delineate the model’s intrinsic capability boundary, reflecting unstable visual perception and reasoning: the policy model may correctly interpret or infer in certain rollouts, yet misperceive the image or generate erroneous deductions in others. Consequently, they constitute ideal training signals for eliciting and strengthening implicit visual re-examination within our self-evolving framework.

\paragraph{Step 2: Information-Gain-Driven Reflection Data Synthesis.}
Based on the pass rate \( P_{\text{pass}}(s) \), 
we partition the trajectory set \( \mathcal{T}(s) \) into three regimes to construct our reflection data.
\begin{align}
\mathcal{T}_{\text{stable}}(s)
&=
\mathcal{T}(s)
\quad \text{if } P_{\text{pass}}(s)=1
\\[6pt]
\mathcal{T}_{\text{unstable}}^{\text{right}}(s)
&=
\{ \tau \in \mathcal{T}(s)
\mid 0 < P_{\text{pass}}(s) < 1,\, 
\tau_{\text{ans}}=a_s
\}
\\[6pt]
\mathcal{T}_{\text{unstable}}^{\text{wrong}}(s)
&=
\{ \tau \in \mathcal{T}(s)
\mid 0 < P_{\text{pass}}(s) < 1,\, 
\tau_{\text{ans}}\neq a_s
\}
\end{align}
To transform above rollout trajectories into structured and learnable 
visual re-examination signals, we standardize the response format 
as illustrated in Fig.~\ref{fig:pipline}. 
This template explicitly separates initial reasoning from reflective 
re-grounding, thereby enforcing an internal two-stage inference process.

Following the standardized template, 
the segment under “think with the image”, 
denoted as \( \tau_{\text{img}} \), 
is extracted from the first-stage reasoning 
of a sampled rollout \( \tau \in \mathcal{T}(s) \). 
To synthesize the \texttt{<reflection>} block, 
we concatenate \( \tau_{\text{img}} \) with the original input 
\( s = (i, q) \) and feed the augmented prompt 
back into the same policy model for a second rollout. 
This design explicitly conditions the model on its own prior 
reasoning trace, enabling deliberate visual re-examination.

To regularize the reflection behavior, 
we impose a strict \textbf{Information Gain} criterion, 
ensuring that the generated reflection introduces 
additional visual evidence, corrective grounding, 
or robustness verification beyond the initial reasoning. 
Specifically, we synthesize three distinct types of reflection steps based on filtered subsets of trajectories:

\begin{enumerate}

\item \textbf{Elaborative Reflection} 
for \( \tau \in \mathcal{T}_{\text{stable}}(s) \):  
when the initial reasoning is consistently correct, 
we encourage broader visual exploration and supplementary 
evidence grounding to reinforce confidence and enhance 
reasoning coherence.

\item \textbf{Corrective Reflection} 
for \( \tau \in \mathcal{T}_{\text{unstable}}^{\text{wrong}}(s) \):  
the reflection explicitly identifies prior visual 
misinterpretations, localizes erroneous regions, 
and re-grounds the reasoning on correct visual evidence 
to repair the inference trajectory.

\item \textbf{Recheck Reflection} 
for \( \tau \in \mathcal{T}_{\text{unstable}}^{\text{right}}(s) \):  
although the initial reasoning yields the correct answer, 
this step enforces deliberate visual re-attention 
to verify that no cues were misperceived, overlooked, 
or spuriously correlated, thereby reducing accidental 
correctness and improving robustness.
\end{enumerate}

Overall, this two-stage formulation encourages the model 
to revisit the image conditioned on its own intermediate 
reasoning, producing a structured and self-corrective 
visual re-examination process.

\paragraph{Step 3: Cold-Start SFT.}

After generating VRE trajectories, 
we retain only those rollouts \( \tau \in \mathcal{T}(s) \) 
whose final prediction satisfies 
\( \tau_{\text{ans}} = a_s \). 
The retained trajectories form a filtered supervision set 
\( \mathcal{D}_{\text{cold}} \). The training objective is as follows:
\begin{equation}
\mathcal{L}_{\text{cold-start}}
=
- \mathbb{E}_{(s,\tau)\sim \mathcal{D}_{\text{cold}}}
\sum_{t=1}^{|\tau|}
\log \pi_\theta ( \tau_t \mid i, q, \tau_{<t} ),
\end{equation}
Then we obtain the cold-start model \( \pi_{\text{SFT}}^{(0)} \), which is endowed with VRE capability.

\subsection{Closed-Loop RLVR and SFT}
\label{sec:model_training}

\subsubsection{RL Data Construction.}

We apply the \( \pi_{\text{SFT}}^{(0)} \) 
to perform dynamic rejection sampling on the training queries. 
For each query, multiple reflection-formatted trajectories are generated, 
and only those passing verifiable outcome checks are retained. 
To focus RL optimization on the model’s cognitive boundary, 
we discard two extremes: intractable queries (hard prompts) 
and already-mastered cases. 
The remaining unsettled queries, in which the model shows partial correctness, reasoning instability, or sensitivity to reflection, form a focused and information-dense RL data pool.
\vspace{-10pt}

\begin{algorithm}[t]
\caption{Reward Computation for GRPO}
\label{alg:reward}
\begin{algorithmic}[1]
\REQUIRE Generated trajectory $\tau$, Ground Truth $a_s$
\ENSURE Total Reward $R(\tau)$

\STATE \textbf{Parse:}
$(\tau_{\text{refl}}, \tau_{\text{ans}}) 
\leftarrow \text{Parse}(\tau)$

\STATE \textbf{Format Reward $R_{\text{form}}$:}
\IF{tags unbalanced OR required blocks missing}
    \STATE $R_{\text{form}} \leftarrow -1$
\ELSIF{$\tau_{\text{refl}}$ empty OR meaningless}
    \STATE $R_{\text{form}} \leftarrow -2$
\ELSIF{$\text{len}(\tau_{\text{ans}}) \ge 1000$}
    \STATE $R_{\text{form}} \leftarrow -1$
\ELSE
    \STATE $R_{\text{form}} \leftarrow 0$
\ENDIF

\STATE \textbf{Reflection Score:}
\STATE $s_{\text{refl}} 
\leftarrow 
\text{LLM}_{\text{scorer}}(\tau)$
\STATE $R_{\text{refl}} 
\leftarrow 
\mathbb{I}(s_{\text{refl}} > 0 
\land \tau_{\text{refl}} \text{ exists})$

\STATE \textbf{Accuracy Score:}
\STATE $R_{\text{acc}} 
\leftarrow 
\text{LLM}_{\text{judge}}(\tau_{\text{ans}}, a_s)$

\STATE \textbf{Combine:}
\STATE $R(\tau) \leftarrow 
\lambda_{\text{form}} R_{\text{form}}
+ \lambda_{\text{acc}} R_{\text{acc}}
+ \lambda_{\text{refl}} R_{\text{refl}}$

\RETURN $R(\tau)$
\end{algorithmic}
\end{algorithm}

\subsubsection{GRPO and Reward Evaluation.}

We optimize the policy on the RL data using Group Relative Policy Optimization (GRPO), 
sampling a group of $G$ trajectories 
$\{\tau_1, \dots, \tau_G\}$ per query. 
The GRPO objective maximizes the relative advantage 
while constraining the KL divergence against the Cold-Start model.

To comprehensively assess the multi-step reasoning trajectory, 
the total reward function $R(\tau_i)$ is formulated as a weighted 
combination of three components:

\begin{equation}
R(\tau_i) =
\lambda_{\text{form}} R_{\text{form}}(\tau_i)
+ \lambda_{\text{acc}} R_{\text{acc}}(\tau_i)
+ \lambda_{\text{refl}} R_{\text{refl}}(\tau_i).
\end{equation}

The coefficients $\lambda_{\text{form}}$, 
$\lambda_{\text{acc}}$, and 
$\lambda_{\text{refl}}$ 
reflect implementation priorities. 
Specifically, we define:
\begin{itemize}
\item \textbf{Format Reward ($R_{\text{form}}$):} 
A \textit{penalty-based} component ($0, -1,$ or $-2$) 
that enforces structural integrity. 
It assigns negative rewards for tag mismatches 
(e.g., unbalanced \texttt{<reflection>} or 
\texttt{<answer>} tags), missing required blocks, 
excessive answer length, 
or meaningless reflection content $\tau_{refl}$ extracted from $\tau$, 
facilitating reliable parsing of the trajectory.

\item \textbf{Accuracy Reward ($R_{\text{acc}}$):} 
A binary reward ($0$ or $1$) determined by an 
\textit{LLM-based judge}. 
The extracted answer content $\tau_{\text{ans}}$ 
is compared against the ground-truth answer $a$ 
to evaluate semantic correctness, 
ensuring robust assessment of the final outcome.

\item \textbf{Reflection Reward ($R_{\text{refl}}$):} 
A binary reward ($0$ or $1$) 
that evaluates the quality of the 
\texttt{<reflection>} block. 
The trajectory $\tau$ is judged using an 
\textit{LLM-based scorer}, 
encouraging reflection that meaningfully builds upon 
the initial visual reasoning (information-gain) rather than empty verbosity.

\end{itemize}

The detailed reward computation procedure 
is presented in Algorithm~\ref{alg:reward}. 
After reinforcement learning, 
we obtain the updated policy 
\( \pi_{\text{RL}}^{(0)} \), which exhibits improved structural compliance, reflection consistency, and answer accuracy.

\subsubsection{Post-RL Self-Distillation}
\label{sec:iterative_evolution}
While the RL model \( \pi_{\text{RL}}^{(0)} \) 
substantially outperforms the preceding cold-start model 
\( \pi_{\text{SFT}}^{(0)} \), we observe a key limitation: simply extending RL training does not yield sustained improvements and may even cause degradation. Prolonged optimization often leads to reward overfitting, training instability, and degradation of general visual capabilities.  

To address this bottleneck, we introduce an iterative refinement phase to close the self-evolution loop. Specifically, the RL model \( \pi_{\text{RL}}^{(0)} \)  revisits previously discarded hard prompts via iterative generation. Benefiting from enhanced visual re-examination ability, it can now solve a subset of these challenging cases. We then apply rejection sampling to retain high-quality reasoning traces, constructing the updated SFT dataset \( \mathcal{D}_{\text{new}} \). Fine-tuning the RL model on \( \mathcal{D}_{\text{new}} \) yields an updated policy \( \pi_{\text{SFT}}^{(1)} \).

\section{Experiments}

\subsection{Experimental Setup}

\paragraph{Implementation Details.} 
To validate our proposed framework, we select Qwen2.5-VL-7B~\cite{bai2025qwen2} as our primary base model for all SFT and RLVR optimization, given its strong open-source baseline performance.  
For SFT, training is conducted on $8\times$ Ascend 910B(64G) NPUs. The RLVR phase scales to $64\times$ Ascend 910B NPUs, employing a 3D parallelism strategy with tensor, pipeline, and data parallelism degrees set to 4, 4, and 4, respectively. Our training infrastructure is built upon LLaMA-Factory for SFT and the Mindspeed-RL framework for reinforcement learning. 
The cold-start data is from Thyme-SFT~\cite{zhang2025thyme}, and the RL training data is from MM-Eureka~\cite{meng2025mmeureka}, $V^*$ dataset~\cite{wu2024v}, and ViRL39K~\cite{wang2025vl}.
Details regarding the training data composition, hyperparameters for the cold-start and GRPO stages, and the judge model prompt are provided in the Appendix.
\begin{table*}[t]
\centering
\caption{\textbf{Performance on Mathematical and Logical Reasoning Benchmarks.} We compare our VRE model with representative closed-source, open-source general, and open-source reasoning MLLMs across mathematical and logical reasoning Benchmarks. $^*$ denotes the results are reproduced by ourselves.}
\vspace{-8pt}

\resizebox{\linewidth}{!}{
\begin{tabular}{lcc|cccccc}
\toprule
\textbf{Model} & \textbf{Tool Use} & \textbf{Param Size} & MathVista & MathVerse & MathVision & WeMath & LogicVista & MMMU \\
\midrule
\rowcolor{gray!15} 
\multicolumn{9}{c}{\textbf{Closed-Source MLLMs}} \\ 
GPT-4o~\cite{achiam2023gpt} & \ding{55} & - & 63.8 & 50.8 & 30.4 & 69.0 & 53.2 & 69.1 \\
Claude 3.7~\cite{claude3.7sonnet} & \ding{55} & - & 66.8 & 52.0 & 41.3 & 72.6 & - & - \\
\addlinespace[0.5em]
\rowcolor{gray!15} 
\multicolumn{9}{c}{\textbf{Open-Source General MLLMs}} \\
LLaVA-OneVision~\cite{li2024llava} & \ding{55} & 7B & 58.6 & 19.3 & 18.3 & 20.9 & 33.3 & 48.8 \\
Qwen2.5-VL~\cite{bai2025qwen2} & \ding{55} & 7B & 68.2 & 49.2 & 23.7$^*$ & 62.1 & 45.9 & 51.2$^*$ \\
InternVL2.5~\cite{chen2024expanding}  & \ding{55} & 8B & 64.4 & 39.5 & 22.0 & - & - & 56.0 \\
InternVL3~\cite{zhu2025internvl3} & \ding{55} & 8B & 71.6 & 39.8 & 29.3 & 37.1 & - & 65.6 \\
\addlinespace[0.5em]
\rowcolor{gray!15}
\multicolumn{9}{c}{\textbf{Open-Source Reasoning MLLMs}} \\
DeepEyesV2~\cite{hong2025deepeyesv2} & \checkmark & 7B & 71.9 & 52.7 & 28.9 & 38.1 & 48.7 & - \\
Thyme~\cite{zhang2025thyme} & \checkmark & 7B & 70.0 & 39.1 & 27.6 & 39.3 & 49.0 & - \\
Semantic-back~\cite{yang2025look} & \ding{55} & 7B & 71.6 & 50.5 & 27.7 & 71.3 & - & 52.0$^*$ \\ 
MM-Eureka~\cite{meng2025mm} & \ding{55} & 8B & 67.1 & 40.0 & 24.7 & 60.1 & - & - \\
R1-VL~\cite{zhang2025r1} & \ding{55} & 8B & 63.5 & 40.4 & 22.2 & 58.7 & - & - \\
\midrule
\rowcolor{tableblue} 
\textbf{VRE (Ours)} & \ding{55} & 7B & \textbf{71.2} & \textbf{53.1} & \textbf{26.5} & \textbf{68.7} & \textbf{48.7} & \textbf{52.1} \\
($\Delta$ vs Qwen2.5-VL 7B) & - & - & \textit{+3.0}& \textit{+3.9}& \textit{+2.8}& \textit{+6.6} & \textit{+2.8} & \textit{+0.9} \\
\bottomrule
\end{tabular}
}
\label{tab:math-reasoning-performance}
\vspace{-15pt}
\end{table*}

\paragraph{Evaluation Benchmarks.} 
To comprehensively evaluate our framework, we select a diverse suite of benchmarks across four crucial dimensions. 
Mathematical Reasoning (MathVerse~\cite{zhang2024mathverse}, MathVision~\cite{wang2024measuring}, MathVista~\cite{lu2023mathvista}, WeMath~\cite{qiao2024we}, LogicVista~\cite{xiao2024logicvista}) tests the model's ability to deduce complex visual math problems in both multiple-choice and free-form formats. 
High-Resolution Perception ($V^*$-Bench~\cite{wu2024v}) measures fine-grained visual grounding and the localization of elusive microscopic details. 
 Visual Document Understanding (ChartQA~\cite{masry2022chartqa}, OCRBench\_v2~\cite{fu2024ocrbench}) validates precise visual data extraction and verification from complex charts and dense text. Finally, Multi-Disciplinary Reasoning (MMMU~\cite{yue2024mmmu}) verifies expert-level generalizability and domain knowledge alignment across college-level disciplines.

\subsection{Main Results}

\begin{table*}[t]
\centering
\caption{\textbf{Performance on High-Resolution Perception, Document Understanding, and Real-world QA.} We compare our VRE model with representative closed-source, open-source general, and open-source reasoning MLLMs across high-resolution perception, document understanding, and Real-world QA. $^*$ denotes the results are reproduced by ourselves.}
\vspace{-8pt}

\resizebox{\linewidth}{!}{
\begin{tabular}{l c c | c c c | c c c | c c | c}
\toprule
\multirow{2}{*}{\textbf{Model}} 
& \multirow{2}{*}{\makecell{\textbf{Tool}\\\textbf{Use}}} 
& \multirow{2}{*}{\makecell{\textbf{Param}\\\textbf{Size}}} 
& \multicolumn{3}{c|}{$V^*$-Bench} 
& \multicolumn{3}{c|}{ChartQA} 
& \multicolumn{2}{c|}{OCRBench\_v2} & \multicolumn{1}{c}{RealworldQA}
\\
\cmidrule(lr){4-6} \cmidrule(lr){7-9} \cmidrule(lr){10-11} \cmidrule(lr){12-12}
& & & Attr & Spatial & Overall & Human & Aug & Overall & en & zh & - \\
\midrule

\rowcolor{gray!15} 
\multicolumn{12}{c}{\textbf{Closed-Source MLLMs}} \\
GPT-4o~\cite{achiam2023gpt} & \ding{55} & - & 60.5 & 67.5 & 79.5 & 91.9 & 85.7 & 86.7 & 46.5 & 32.2 & 75.5 \\

\addlinespace[0.5em]
\rowcolor{gray!15} 
\multicolumn{12}{c}{\textbf{Open-Source General MLLMs}} \\
LLaVA-OneVision~\cite{li2024llava} & \ding{55} & 7B  & 75.7 & 75.0 & 75.4 & - & - & 80.0 & - & - & 66.3 \\
InternVL2.5~\cite{chen2024expanding} & \ding{55} & 8B & - & - & - & - & - & 84.8 & 49.8 & 52.1 & 70.1 \\
Qwen2.5-VL~\cite{bai2025qwen2} & \ding{55} & 7B  & 78.2$^*$ & 73.6$^*$ & 76.4$^*$ & 72.5 & 94.9 & 83.7 & 56.3 & 57.2 & 68.2 \\
Qwen2.5-VL~\cite{bai2025qwen2} & \ding{55} & 32B & 77.4$^*$ & 86.8$^*$ & 81.2$^*$ & 76.9 & 82.6 & 81.1 & - & - & 70.2 \\
\addlinespace[0.5em]
\rowcolor{gray!15}
\multicolumn{12}{c}{\textbf{Open-Source Reasoning MLLMs}} \\
SEAL~\cite{wu2024v} & \checkmark & 7B  & 74.8 & 76.3 & 75.4 & - & - & - & - & - & - \\
DyFo~\cite{li2025dyfo} & \checkmark & 7B  & 80.0 & 82.9 & 81.2 & - & - & - & - & - & - \\
DeepEyesV2~\cite{hong2025deepeyesv2} & \checkmark & 7B  & - & - & 81.8 & - & - & 88.4 & - & - & - \\
Thyme~\cite{zhang2025thyme} & \checkmark & 7B  & 83.5 & 80.3 & 82.2 & 80.0 & 92.2 & 86.1 & - & - & 70.2 \\
Semantic-back~\cite{yang2025look} & \ding{55} & 7B  & 80.2$^*$ & 78.7$^*$ & 79.6$^*$ & - & - & - & - & - & - \\
\midrule
\rowcolor{tableblue} 
\textbf{VRE (Ours)} & \ding{55} & 7B  & \textbf{84.3} & \textbf{82.9} & \textbf{83.8} & \textbf{87.3} & 90.4 & \textbf{88.8} & \textbf{62.6} & \textbf{64.7} & 69.8 \\
($\Delta$ vs Qwen2.5-VL 7B) & - & - & \textit{+6.1}& \textit{+9.3}& \textit{+7.4}& \textit{+14.8} & \textit{-4.5} & \textit{+5.1} & \textit{+6.3} & \textit{+7.5} & \textit{+1.6} \\
\bottomrule
\end{tabular}
}
\label{tab:comprehensive-performance}
\vspace{-15pt}
\end{table*}

We evaluate VRE across four major domains: Mathematical and Multi-Disciplinary Reasoning (Table~\ref{tab:math-reasoning-performance}), and High-Resolution Perception and Structured Visual Understanding (Table~\ref{tab:comprehensive-performance}). 

A striking observation across our evaluations is that VRE achieves high performance among 7B-scale models without relying on explicit tool invocation or architectural modifications. For instance, on the High-Resolution $V^*$-Bench (Table~\ref{tab:comprehensive-performance}), VRE yields an impressive Overall score of 83.8\%, completely outperforming tool-augmented reasoning models such as Thyme (82.2\%), DeepEyesV2 (81.8\%), and DyFo (81.2\%). Furthermore, compared to the base model Qwen2.5-VL-7B, VRE delivers substantial absolute gains across perception and extraction tasks, including +7.4\% on $V^*$-Bench, +5.1\% on ChartQA, and up to +7.5\% on OCRBench\_v2. This robust improvement confirms that implicit visual re-examination natively solves the visual drift problem without external cropping or zooming tools.

As illustrated in Table~\ref{tab:math-reasoning-performance}, VRE exhibits exceptional capabilities in deep logical tracking and mathematical deduction.
Specifically, VRE achieves 71.2 on MathVista and 53.1 on MathVerse, securing absolute gains of +3.0\% and +3.9\% over the base model, respectively. This consistent trajectory of improvement extends across other rigorous benchmarks, yielding a striking +6.6\% gain on WeMath (68.7) and solid enhancements on MathVision (+2.8\%), LogicVista (+2.8\%), and MMMU (+0.9\%). 
Crucially, despite operating with only 7 billion parameters, VRE significantly narrows, and in several cases completely bridges, the performance gap with massive closed-source proprietary systems. This parameter-efficient capability empirically validates that our self-iterative reflection mechanism effectively unlocks deep cognitive reasoning without relying on brute-force parameter expansion.

\subsection{Mechanism Analysis and Ablation Studies}

\subsubsection{Component Breakdown and Evolutionary Dynamics.} 
To validate the necessity of our self-iterative pipeline, we conduct a rigorous ablation study on major benchmarks.

\begin{table}[t]
\centering
\caption{\textbf{Component Ablation.} Performance on select benchmarks when progressively adding VRE components.}
\label{tab:ablation_components}
\vspace{-8pt}

\resizebox{\linewidth}{!}{
\begin{tabular}{l c c c | c c c c c c}
\toprule
Setting & Cold-SFT & RLVR & Iter-SFT & $V^*$-Bench & MathVista & MathVerse & ChartQA & OCRBench\_v2\textsubscript{(en/zh)} \\
\midrule
Base Model & \ding{55} & \ding{55} & \ding{55} & 76.4 & 68.2 & 49.2 & 83.7 & 56.3/57.2\\
SFT Only   & \checkmark & \ding{55} & \ding{55} & 80.1 & 69.2 & 49.6 & 87.8 & 59.6 / 62.3 \\
SFT + RL   & \checkmark & \checkmark & \ding{55} & 83.8 & 71.0 & 52.5 & 88.7 & 62.1 / 64.3\\ 
\rowcolor{tableblue}
\textbf{VRE} & \checkmark & \checkmark & \checkmark & 84.3 & 71.2 & 53.1 & 88.8 & 62.6 / 64.7\\
\bottomrule
\end{tabular}
}
\vspace{-12pt}
\end{table}
\paragraph{Efficacy of Cold-Start SFT: Activating Visual Grounding.}
As shown in the second row of Table~\ref{tab:ablation_components}, SFT Only yields immediate and substantial gains on perception-heavy and structured understanding tasks. This indicates that supervised fine-tuning successfully establishes the format prior for the \texttt{<reflection>} template, allowing the model to consciously trigger visual re-examination for data extraction. However, the gain on complex reasoning tasks is marginal (e.g., MathVerse only improves by +0.4\%), suggesting that SFT alone is insufficient to correct deep logical errors.

\paragraph{Impact of RLVR: Unlocking Deep Reasoning.}
As shown in the third row of Table~\ref{tab:ablation_components}, the integration of SFT + RLVR catalyzes a critical breakthrough in reasoning capabilities. Most notably, MathVerse witnesses a remarkable leap of \textit{\textit{+2.9\%}} (49.6 $\rightarrow$ 52.5), and $V^*$-Bench further climbs to 83.8. This confirms that our Reflection Rewards ($R_{\text{refl}}$) are essential drivers. They transform the model's behavior from merely mimicking the reflection format to actively utilizing visual evidence for logical self-correction, thereby solving complex problems that SFT could not handle.

\paragraph{Necessity of Post-RL SFT: Stabilizing the Policy.}
Finally, As shown in the fourth row of Table~\ref{tab:ablation_components}, the full VRE pipeline provides a consistent stability boost across all benchmarks. By salvaging ``hard prompts'' that were initially unstable during RL exploration, VRE pushes the performance ceiling further (e.g., MathVerse reaches 53.1, MathVista reaches 71.2). While the magnitude of these gains is smaller compared to the RL stage, their universality across all domains proves that the self-iterative loop effectively consolidates the policy and expands the model's cognitive boundary into the long-tail distribution of difficulty.
More example analyses can be found in the Appendix.

\begin{figure}[t]
\centering
\includegraphics[width=1\linewidth]{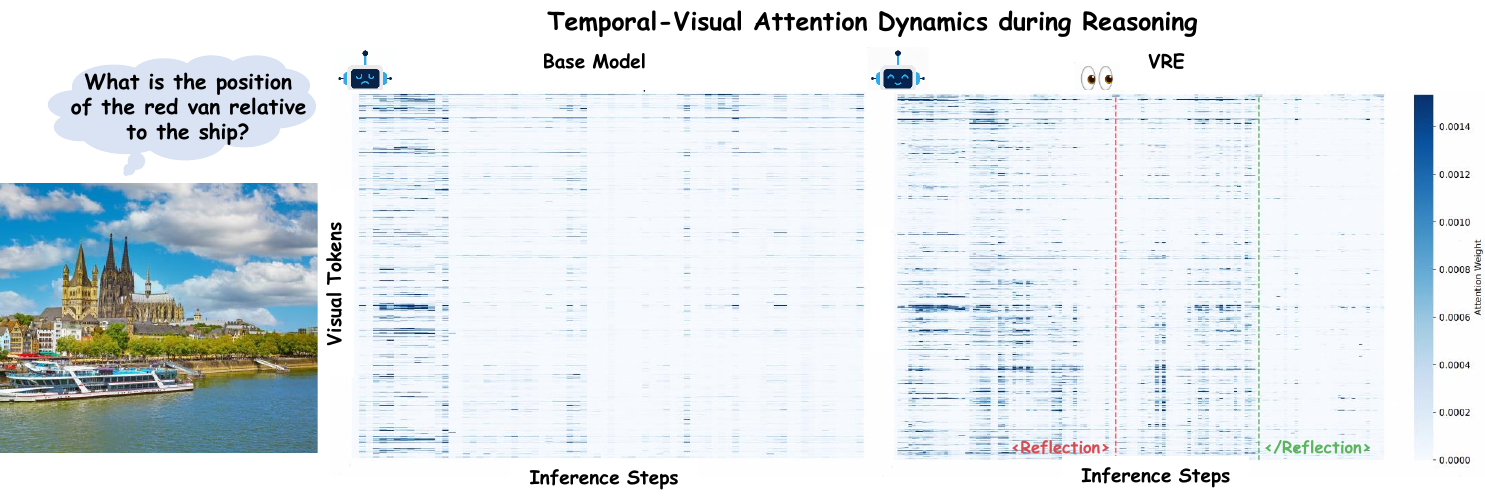}
\vspace{-18pt}
\caption{\textbf{Visualizing Implicit Visual Re-Examination via Attention.} 
The heatmaps display attention weights on visual tokens across inferences steps. 
\textbf{(Left)} The Base Model exhibits visual decay, where attention to the image vanishes as the textual context grows, leading to ungrounded hallucinations. 
\textbf{(Right)} The VRE Model demonstrates a spontaneous attention resurgence. Inside the \texttt{<reflection>} block, the model sharply re-allocates attention back to the visual features, proving that the mechanism actively triggers a re-examination behavior to extract missing visual evidence.}
\label{fig:visual_m}
\vspace{-10pt}
\end{figure}

\subsubsection{Anatomy of Visual Re-Examination.}\paragraph{Visual Attention Re-allocation.} 
To verify that VRE genuinely reactivates visual processing rather than hallucinating based on text, we visualize the average attention weights over visual tokens across decoding steps. 
As shown in Figure~\ref{fig:visual_m}, the base model suffers from severe visual decay: as the reasoning chain lengthens, the attention to image features monotonically fades, indicating a drift toward pure textual reliance. 
In sharp contrast, VRE exhibits a distinct visual attention resurgence. Specifically, upon careful re‑examination of the content, within the generated \texttt{<reflection>} block, the attention weights on visual tokens spike significantly.
This phenomenon confirms that VRE successfully reactivates the dormant visual encoder to re-examine evidence, effectively reversing the visual drift and grounding the reasoning back to the image.

\begin{figure}[t]
\centering
\includegraphics[width=1\linewidth]{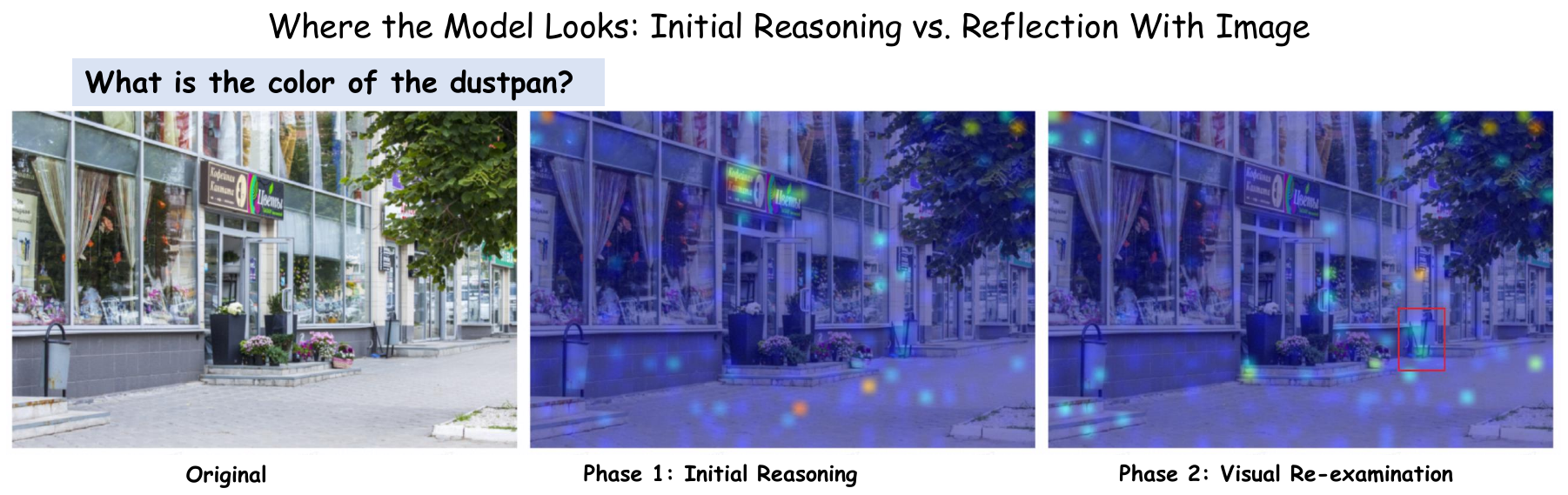}
\vspace{-20pt}
\caption{\textbf{Visualizing the Re-examination Mechanism: From Blindness to Grounding.}
Phase 1: Initial Reasoning.  During the initial pass, the model's attention is dispersed over the background, failing to locate the dustpan. 
Phase 2: Visual Re-examination. Triggered by the reflection token, the model performs an active visual search. The attention map shows a sharp, targeted focus on the dustpan (red region), successfully retrieving the correct visual evidence to fix the answer. 
}
\label{fig:visual_focusing}
\vspace{-8pt}
\end{figure}

\paragraph{Introspective Visual Re-Examination.} 
While the attention matrix confirms the temporal return of visual signals, we further investigate the spatial mechanism of this re-examination: What exactly does the model look at when the visual signal returns? 
Figure~\ref{fig:visual_focusing} provides a qualitative visualization of the spatial attention shift.
In the Initial Reasoning (Phase 1), the model exhibits typical ``visual blindness,'' where its attention drifts to irrelevant background textures, failing to ground the target object. Lacking visual evidence, the model hallucinates a color based on textual priors.
However, during the Reflection (Phase 2), the VRE mechanism triggers a targeted Corrective Re-grounding. As shown in the heatmap, the attention spotlight does not just randomly increase; it performs a precise search-and-lock operation, shifting specifically to the dustpan's spatial coordinates.
This qualitative evidence completes our mechanistic understanding: the re-examination process works by (1) identifying missing information (When), (2) spatially re-grounding the specific entity (Where), and (3) extracting the correct attribute to overwrite the hallucination (Why).

\paragraph{Evolution of Reflection Paradigms.} 
To deeply understand how the VRE framework guides the model to master implicit visual re-examination, we track the fine-grained distribution of reflection paradigms across different training stages. As defined in Sec.~\ref{sec:coldstart}, generated reflections are categorized into three valid Information Gain types, alongside a No Gain category representing redundant self-paraphrasing. To quantify this evolution, we randomly sample 100 test trajectories per benchmark to analyze the proportion of these four categories.

\begin{figure*}[t]
    \centering
    \includegraphics[width=\linewidth]{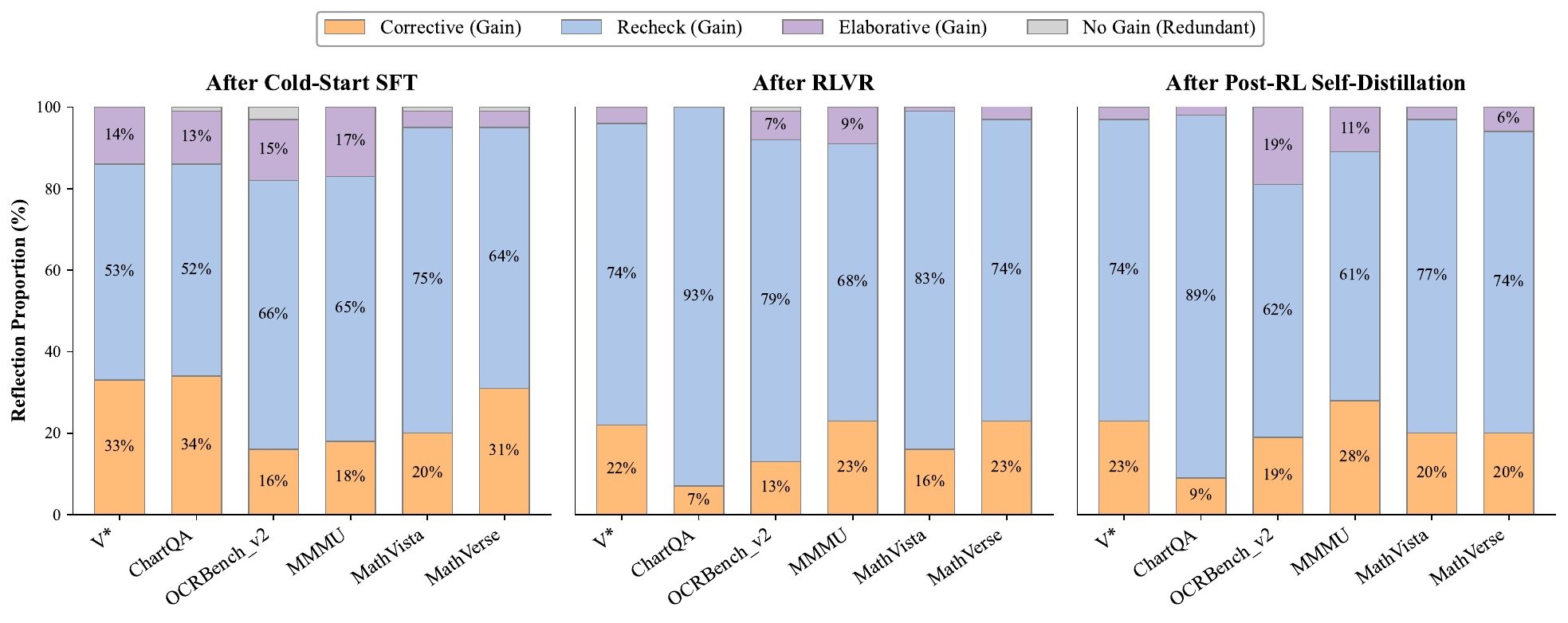} 
    \vspace{-20pt}
    \caption{\textbf{Evolution of Reflection Paradigms.} The stacked bar charts detail the distribution of introspection types across training stages.}
    \label{fig:reflection_distribution}
\vspace{-10pt}
\end{figure*}

Figure~\ref{fig:reflection_distribution} illustrates this evolutionary distribution as stacked bar charts across six diverse benchmarks. After the initial Cold-Start SFT, the model exhibits merely a superficial understanding of introspection, generating No Gain reflections relatively frequently. This confirms that explicitly mimicking the <reflection> template does not guarantee genuine visual grounding. However, after the RLVR and the subsequent Post-RL Self-Distillation, the proportion of No Gain drops drastically, approaching zero. This decisively validates our Reflection Reward, which successfully penalizes empty verbosity and enforces tangible visual evidence extraction.
First, the frequency of Corrective reflections is notably higher in tasks like $V^*$, MMMU, and MathVerse. This aligns perfectly with their inherent challenges: $V^*$ requires locating elusive visual details easily missed during initial perception, while MMMU involves deep, complex logical chains. In these scenarios, the model intelligently learns to use the image to debug and rectify its early visual misinterpretations or flawed reasoning paths. 
Second, Recheck behaviors are distinctly prominent in mathematical and chart-based benchmarks. Since these tasks heavily rely on precise data extraction, the model learns to execute a final, rigorous visual verification step before outputting the answer to prevent careless hallucinations. 
Conversely, Elaborative reflections see the highest ratio in OCR-heavy and multi-disciplinary contexts. For tasks requiring dense text reading and rich contextual gathering, the model naturally favors explicitly augmenting its reasoning with supplementary visual transcriptions and descriptive details to build a solid premise. 
Ultimately, these fine-grained structural adaptations firmly prove that VRE cultivates a genuinely intelligent, self-directed visual introspection mechanism that dynamically tailors its cognitive strategy to the specific difficulty of diverse multimodal tasks.

\subsubsection{Statistical Verification of the Reflection Mechanism.}
\label{sec:reflction_ana}
Beyond macroscopic benchmark improvements, we employ information-theoretic metrics to rigorously validate two fundamental properties of our VRE framework: the training stability of same policy, and the genuine utility of the generated visual reflections. Experimental results can be found in the Appendix.

\section{Conclusion}
\label{sec:conclusion}

In this paper, we address the ``visual drift'' phenomenon in long-form multimodal generation, where models neglect visual evidence for textual priors. We propose Visual Re-Examination (VRE), a self-iterative framework that elicits implicit visual refocusing without expensive interventions. By integrating a Reflection Information Gain metric into a continual loop spanning data curation and RLVR optimization, VRE ensures targeted visual evidence retrieval to correct reasoning trajectories. Evaluations demonstrate competitive performance among 7B-scale models without requiring architectural changes or explicit tool use.



%
%
\appendix

\section{Implementation Details}
\subsection{Data Resources}
\label{sec:appendix_data}

We systematically curate a high-quality data composition specifically tailored for our three-stage pipeline: (1) Cold-Start SFT for embedding visual reflection priors, (2) RLVR optimization via GRPO, and (3) Post-RL Self-Distillation. The detailed dataset sources and quantities are summarized in Table~\ref{tab:data_resources}.

\begin{table}[h]
\centering
\vspace{-15pt}
\caption{\textbf{Detailed Composition of Training Data.} This table summarizes the data sources and the approximate number of samples in our VRE Framework.}
\label{tab:data_resources}
\resizebox{0.8\linewidth}{!}{
\begin{tabular}{l |c|  c | c}
\toprule
\multirow{1}{*}{\textbf{Data Source}} 
 & \textbf{Cold-Start SFT} & \textbf{RLVR} & \textbf{Post-RL SFT }\\
\midrule
Thyme-SFT~\cite{zhang2025thyme}   & $\sim$ 80k & - & - \\
MM-Eureka~\cite{meng2025mmeureka} & - & $\sim$ 20k & $\sim$ 4k \\
$V^*$ Dataset~\cite{wu2024v}      & - & $\sim$ 20k & $\sim$ 4k \\
ViRL39K~\cite{wang2025vl}         & - & $\sim$ 20k & $\sim$ 4k \\
\midrule
\rowcolor{gray!10}
\textit{Total Data}        & \textit{$\sim$ 80k} & \textit{$\sim$ 60k} & \textit{$\sim$ 12k} \\
\bottomrule
\end{tabular}
}
\vspace{-15pt}
\end{table}

\subsection{Hyperparameters and Configuration}
\label{sec:appendix_hyperparameters}

\begin{table}[h]
\vspace{-15pt}
\centering
\caption{\textbf{Hyperparameters for SFT and Inference.} 
}
\label{tab:sft_hyperparameters}
\resizebox{\linewidth}{!}{
\begin{tabular}{l|cc|lc}
\hline
\multicolumn{3}{c|}{\textbf{SFT Hyperparameters}} & \multicolumn{2}{c}{\textbf{Inference / Rollout Hyperparameters}} \\
\midrule
\textbf{Parameter} & \textbf{Cold-Start} & \textbf{Post-RL} & \textbf{Parameter} & \textbf{Value} \\
\midrule
Global Batch Size & 64 & 64 & Max New Tokens & 4900 \\
Learning Rate & 1e-5 & 5e-7 & Temperature & 1.0 \\
LR Scheduler & Cosine & Cosine & Top-$p$ & 1.0 \\
Max Sequence Len & 4900 & 4900  & Top-$k$ & -1 \\
Num Epochs & 3 & 1 & Rollout Trajectories ($N$) & 8 
\\
\bottomrule
\end{tabular}
}
\vspace{-15pt}
\end{table}

We detail the hyperparameter configurations for SFT and RLVR in Tables~\ref{tab:sft_hyperparameters} and~\ref{tab:rl_hyperparameters}, respectively. During trajectory rollouts, we fix the sampling count to $N=8$ to balance diversity and cost. For the RLVR stage, we utilize GRPO with strict KL divergence constraints. 
Both the $\text{LLM}_{\text{judge}}$ and $\text{LLM}_{\text{scorer}}$ utilized during the RLVR phase are instantiated using the Qwen2.5-72B~\cite{qwen2025qwen25technicalreport}.
For the text generation during RL, we configure the maximum response length to 7168 tokens and set the temperature to 1.0 to encourage extensive exploration of reasoning trajectories. For the RL configuration, we set $\gamma=1.0$, $\lambda=0.95$, and an initial KL coefficient of 0.01. The GRPO group size is set to $G=10$, with reward coefficients balanced to prioritize reasoning correctness while penalizing redundant reflections. 
\begin{table}[h]
\centering
\vspace{-15pt}
\caption{Hyperparameter configurations for RLVR.}
\label{tab:rl_hyperparameters}
\begin{tabular}{l|l}
\hline

\textbf{Category} & \textbf{Configuration} \\ \hline
Generation & max\_tokens=7168, min\_tokens=50, temp=1.0, top\_p=1.0 \\
RL Algorithm & $\gamma=1.0$, $\lambda=0.95$, adv\_est=group\_norm \\
KL Control & type=fixed, init\_coef=0.01, penalty=kl \\
GRPO & clip\_ratio=0.2, group\_size=10, $\lambda_{\text{form}}$=0.4, $\lambda_{\text{acc}}$=0.6, $\lambda_{\text{refl}}$=0.4 \\
\hline
\end{tabular}
\vspace{-15pt}
\end{table}
\vspace{-15pt}
\subsection{Prompts}
We provide templates tailored for both SFT and RLVR settings to ensure consistent output structures and reliable activation of the <reflection> mechanism. These templates enforce a structured output format (see Figs. \ref{fig:sys_prompt_reflection}, \ref{fig:prompt_example}, \ref{fig:elaborative_prompt_reflection}, \ref{fig:reflection_prompt}, \ref{fig:correct_prompt_reflection} and \ref{fig:recheck_prompt_reflection}).

\vspace{-10pt}
\begin{figure}[htbp]
    \centering
    \begin{tcolorbox}[
        title=\textbf{System Prompt Template for Information Gains},
        colback=white,
        colframe=black,
        fonttitle=\bfseries,
        boxrule=0.8pt,
        arc=2mm,
        boxsep=2pt,          
        left=3pt,            
        right=3pt,           
        top=0pt,             
        bottom=0pt,          
        titlerule=0.5pt,     
        title filled=true,  
    ]
    \begin{lstlisting}[style=promptstyle]
You are an analytical assistant that solves visual reasoning problems through a structured, iterative process focused on maximizing information gain from the image.
    Your reasoning must follow two stages:

    Stage 1: Iterative Reasoning with Visual Information Gain
        - Begin with a <reflection> block: critically examine the initial reasoning and actively seek new visual information by mentally re-examining the image.
        - Ask: "What visual details did I miss?"
        - Focus on overlooked elements: objects, text, colors, spatial relationships, actions, or contextual cues.
        - Each reflection should re-perceive the image to extract additional relevant visual content.
        - Follow thinking with the reflection: reason step by step based on current understanding of the image.
        - The thinking-reflection cycle ought to be information-dense: brief but introducing new image-derived insights that advance the solution.
        - For determining relative positions, please define the positional relationships from the observer's viewpoint.

    Stage 2: Answer
        - When confident, output your final answer inside <answer>...</answer>.
        - No explanations outside the specified tags.

    Key principle: Maximize information density-each reflection should be concise but yield substantial new visual insights.
    \end{lstlisting}
    \end{tcolorbox}
    \caption{System prompt template for information gains. 
    }
    \label{fig:sys_prompt_reflection}
    \vspace{-15pt}
\end{figure}

\begin{figure}[htbp]
    \centering
    \begin{tcolorbox}[
        title=\textbf{Prompt Template for Accuracy Evaluation},
        colback=white,
        colframe=black,
        fonttitle=\bfseries,
        boxrule=0.8pt,
        arc=2mm,
        boxsep=2pt,          
        left=3pt,            
        right=3pt,           
        top=0pt,             
        bottom=0pt,          
        titlerule=0.5pt,     
        title filled=true,  
    ]
    \begin{lstlisting}[style=promptstyle]
Below are two answers to a question. Question is [], 
[Standard Answer] is the standard answer..., and [Model_answer] 
is the answer extracted from a model's output...
Determine whether these two answers are consistent.
If they are consistent, Judgement is 1; if different, Judgement is 0.
Just output Judgement and don't output anything else.

[Question]: Is the countertop tan or blue?
[Standard Answer]: The countertop is tan.
[Model_answer]: tan
Judgement: 1

[Question]: On which side of the picture is the barrier?
[Standard Answer]: The barrier is on the left side of the picture.
[Model_answer]: left
Judgement: 1

... (5 few-shot examples omitted for brevity) ...

[Question]: What color is the towel in the center of the picture?
[Standard Answer]: The towel in the center of the picture is blue.
[Model_answer]: The towel in the center of the picture is pink.
Judgement: 0

[Question]: \promptvar{question}
[Standard Answer]: \promptvar{ground_truth}
[Model_answer]: \promptvar{predict_str}
Judgement:
    \end{lstlisting}
    \end{tcolorbox}
    \caption{Final accuracy evaluation prompt. 
    Dynamic inputs 
    \textcolor{varcolor}{\texttt{\{question\}}} (Question), \textcolor{varcolor}{\texttt{\{ground\_truth\}}} (Standard Answer) and \textcolor{varcolor}{\texttt{\{predict\_str\}}} (Model\_answer) are inserted at inference time.}
    \label{fig:prompt_example}
\end{figure}

\begin{figure}[htbp]
    \centering
    \vspace{-15pt}
    \begin{tcolorbox}[
        title=\textbf{Prompt Template for Elaborative Reflection},
        colback=white,
        colframe=black,
        fonttitle=\bfseries,
        boxrule=0.8pt,
        arc=2mm,
        boxsep=2pt,          
        left=3pt,            
        right=3pt,           
        top=0pt,             
        bottom=0pt,          
        titlerule=0.5pt,     
        title filled=true,  
    ]
    \begin{lstlisting}[style=promptstyle]
The question: {question}. The initial reasoning: {initial_reasoning}. 
You need to re-examine the image to enrich the reasoning with supplementary visual evidence before concluding.

Start with <reflection>, then think with the reflection. 

In each <reflection>:
   - Scan the image for supporting details that were omitted in the initial step.
   - Elaborate on fine-grained attributes: specific colors, textures, small objects, or background context.
   - Explicitly connect these new visual details to strengthen the current reasoning path.
   - Goal: Maximize information density without changing the logical direction.

When ready, provide your definitive final answer inside <answer>...</answer>.
    \end{lstlisting}
    \end{tcolorbox}
    \caption{Prompt Template for Elaborative Reflection. Dynamic inputs 
    \textcolor{varcolor}{\texttt{\{question\}}} (Question), \textcolor{varcolor}{\texttt{\{initial\_reasoning\}}} (Initial Reasoning) are inserted at inference time. 
    }
    \label{fig:elaborative_prompt_reflection}
\end{figure}

\begin{figure}[htbp]
    \centering
    \begin{tcolorbox}[
        title=\textbf{Prompt Template for Reflection Evaluation},
        colback=white,
        colframe=black,
        fonttitle=\bfseries,
        boxrule=0.8pt,
        arc=2mm,
        boxsep=2pt,          
        left=3pt,            
        right=3pt,           
        top=0pt,             
        bottom=0pt,          
        titlerule=0.5pt,     
        title filled=true,  
    ]
    \begin{lstlisting}[style=promptstyle]
You are an expert in reasoning analysis. Your task is to evaluate 
a provided thinking process to determine whether its reflections 
demonstrate a genuine attempt to **gain more visual information 
from the input image by re-examining or re-perceiving its content**.

The thinking process consists of sequential steps: each step of 
reasoning is followed by a <reflection>...</reflection> block.

**Important Context**: 
- The original problem involves an **image** (though you, as a text-only model, cannot see it). 
- The thinker is expected to reason about that image. 
- A valuable reflection should show an **attempt to look back at 
  the image** and extract **additional or previously overlooked 
  visual details** relevant to answering the question.

Evaluate each <reflection>...</reflection> block using these criteria:
1. **Visual Re-examination Intent**: Does the reflection suggest 
   re-inspection of the image to uncover new visual elements 
   (objects, spatial relations, colors, text, etc.)?
2. **Relevant Information Gain**: Does this re-examination lead to **new, image-derived information** that meaningfully advances the reasoning?
Only count a reflection if it satisfies **both** criteria.

Output: a single JSON object with key "score" (integer count).

#### Question:
\promptvar{question}

#### The Thinking Process:
\promptvar{predict_str}

### Output Format (Strictly Enforced)
Return only a valid JSON object-no explanations, no extra text.
Example: {"score": 2}

Your Evaluation Result:
    \end{lstlisting}
    \end{tcolorbox}
    \vspace{-10pt}
    \caption{Prompt for evaluating whether reflection blocks demonstrate 
    genuine visual re-examination. Dynamic inputs 
    \textcolor{varcolor}{\texttt{\{question\}}} (Question) and 
    \textcolor{varcolor}{\texttt{\{predict\_str\}}} (Model\_answer) are inserted at inference time.}
    \label{fig:reflection_prompt}
    \vspace{-5pt}
\end{figure}


\begin{figure}[htbp]
    \centering
    \begin{tcolorbox}[
        title=\textbf{Prompt Template for Corrective Reflection},
        colback=white,
        colframe=black,
        fonttitle=\bfseries,
        boxrule=0.8pt,
        arc=2mm,
        boxsep=2pt,          
        left=3pt,            
        right=3pt,           
        top=0pt,             
        bottom=0pt,          
        titlerule=0.5pt,     
        title filled=true,  
    ]
    \begin{lstlisting}[style=promptstyle]
The question: {question}. The initial reasoning: {initial_reasoning}. 
The initial reasoning may contain visual hallucinations or logical errors. You must critically re-examine the image to detect and rectify them.

Start with <reflection>, then think with the reflection. 

In each <reflection>:
    - Act as a critic: Compare the text strictly against the image pixels.
    - Identify the exact point where the reasoning deviates from the visual reality (the "visual drift").
    - Explicitly state the error (e.g., The text says X, but the image shows Y)
    - Re-ground the reasoning using the correct visual evidence.

When ready, provide your corrected final answer inside <answer>...</answer>.
    \end{lstlisting}
    \end{tcolorbox}
    \vspace{-10pt}
    \caption{Prompt Template for Corrective Reflection. Dynamic inputs 
    \textcolor{varcolor}{\texttt{\{question\}}} (Question), \textcolor{varcolor}{\texttt{\{initial\_reasoning\}}} (Initial Reasoning) are inserted at inference time.  
    }
    \label{fig:correct_prompt_reflection}
\end{figure}

\begin{figure}[htbp]
    \centering
    \begin{tcolorbox}[
        title=\textbf{Prompt Template for Recheck Reflection },
        colback=white,
        colframe=black,
        fonttitle=\bfseries,
        boxrule=0.8pt,
        arc=2mm,
        boxsep=2pt,          
        left=3pt,            
        right=3pt,           
        top=0pt,             
        bottom=0pt,          
        titlerule=0.5pt,     
        title filled=true,  
    ]
    \begin{lstlisting}[style=promptstyle]
The question: {question}. The initial reasoning: {initial_reasoning}. 
You need to perform a rigorous visual audit to double-check the key evidence before finalizing the answer.

Start with <reflection>, then think with the reflection. 

In each <reflection>:
    - Verify precise data points: check specific numbers, text transcription (OCR), or object counts.
    - Re-confirm spatial relationships (left/right, above/below) mentioned in the reasoning.
    - Ensure no key condition in the question was overlooked.
    - Goal: Eliminate carelessness and ensure absolute precision.

When ready, provide your verified final answer inside <answer>...</answer>.
    \end{lstlisting}
    \end{tcolorbox}
    \caption{Prompt Template for Recheck Reflection. Dynamic inputs 
    \textcolor{varcolor}{\texttt{\{question\}}} (Question), \textcolor{varcolor}{\texttt{\{initial\_reasoning\}}} (Initial Reasoning) are inserted at inference time. 
    }
    \label{fig:recheck_prompt_reflection}
    \vspace{-10pt}
\end{figure}

\section{Additional Experiments}

\subsection{Scaling Laws for SFT and RL}
\begin{figure}[htbp]
    \vspace{-5pt}
    \centering
    \begin{subfigure}[b]{0.49\linewidth}
        \centering
        \includegraphics[width=\textwidth]{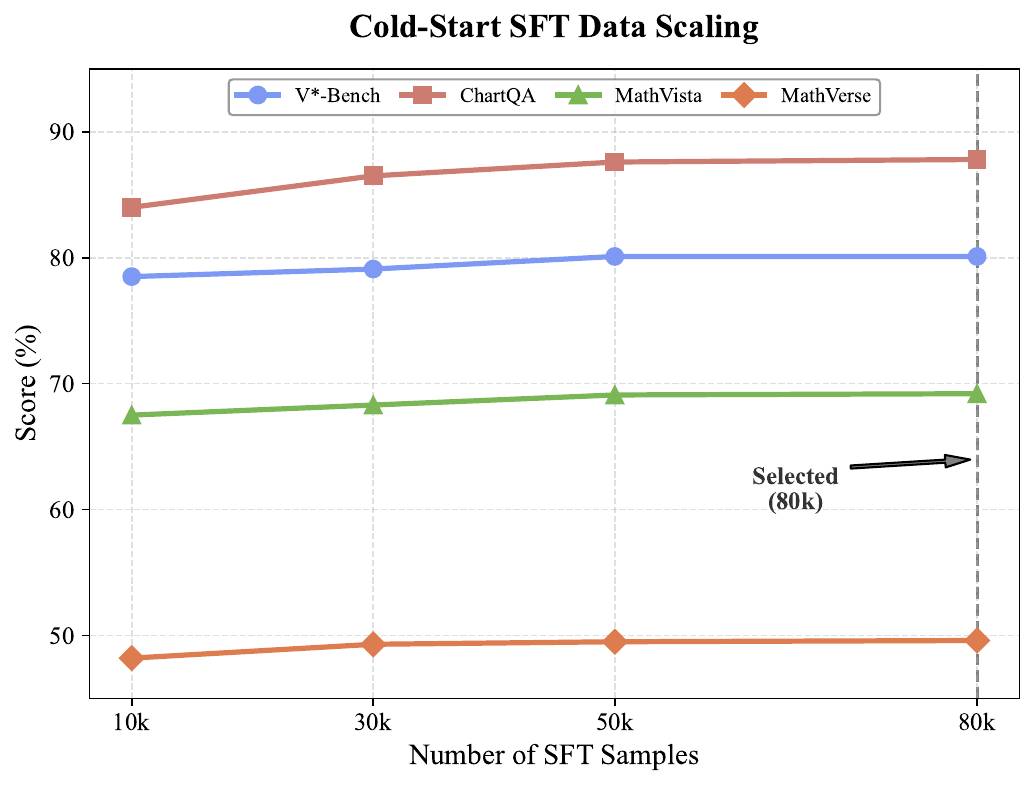}
        \caption{\textbf{Cold-Start SFT Data Scaling}}
        \label{fig:scaling_low_a}
    \end{subfigure}
    \hfill
    \begin{subfigure}[b]{0.49\linewidth}
        \centering
        \includegraphics[width=\textwidth]{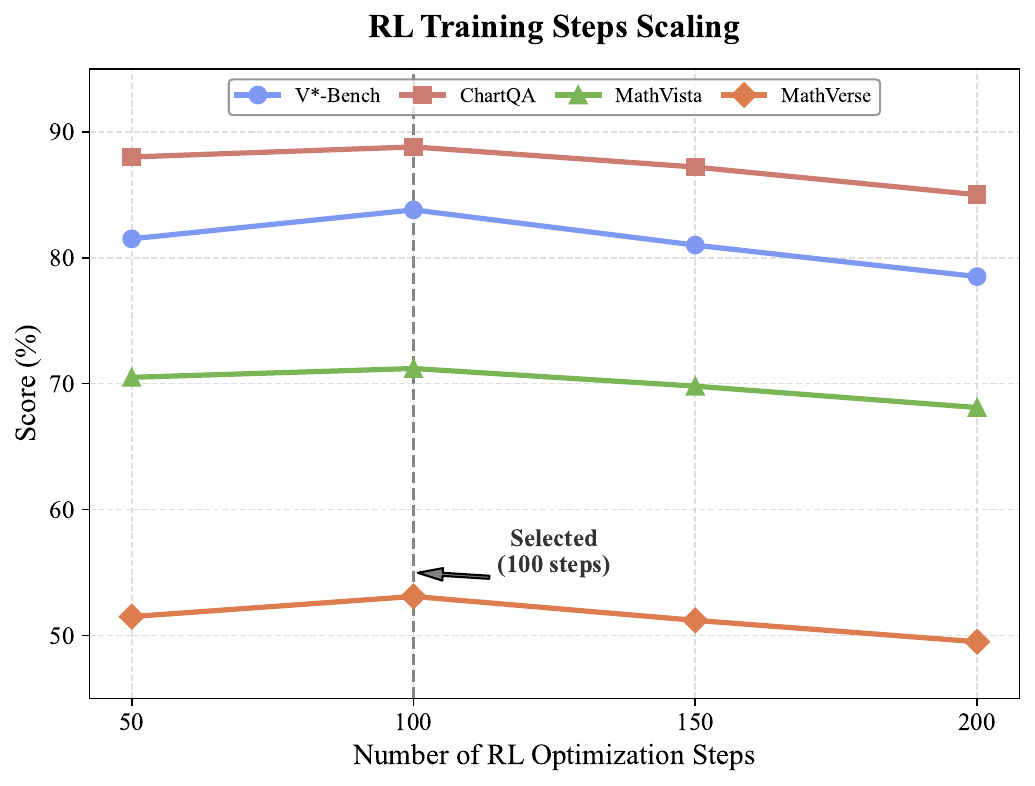}
        \caption{\textbf{RL Training Step Scaling}}
        \label{fig:scaling_low_b}
    \end{subfigure}
    
    \caption{\textbf{Scaling Laws in VRE Training.} 
    \textbf{(a)} Cold-Start SFT Data Scaling. There is no significant performance improvement after 50k samples, and the performance is approaching saturation.
    \textbf{(b)} RL Training Step Scaling. The policy achieves optimal reasoning performance at 100 steps. Prolonged training ($>150$ steps) induces severe over-optimization and capability degradation.}
    \label{fig:scaling_low}
\end{figure}

To deeply understand the optimization bounds of our VRE framework, we systematically ablate the data scale for cold-start SFT and the training steps for RLVR. 
\vspace{-5pt}
\paragraph{Cold-Start SFT Data Scaling.} 
As shown in Figure~\ref{fig:scaling_low_a}, we train the base model with varying quantities of curated cold-start SFT reflection data ($10k$, $30k$, $50k$, and $80k$). The performance exhibits a logarithmic scaling law: accuracy improves rapidly as data scales up to $50k$, demonstrating that the model efficiently internalizes the reflection format prior. However, the marginal gains significantly diminish between $50k$ and $80k$ (e.g., MathVista only increases by +0.1\%). Therefore, we select 80k as the optimal SFT capacity to fully saturate the format alignment without incurring redundant training costs.
\vspace{-5pt}

\paragraph{RL Step Scaling.} 
Figure~\ref{fig:scaling_low_b} tracks the RLVR performance at intervals of 50, 100, 150, and 200 steps. The performance peaks around 100 steps, marking the optimal point where the policy successfully masters information-gain extraction. Crucially, forcing the optimization beyond 150 steps leads to a dramatic performance collapse. This empirical degradation perfectly illustrates the phenomenon of over-optimization (reward hacking). This finding decisively justifies our strategy to halt RL at 100 steps and rely on Post-RL SFT to stabilize the policy.
\vspace{-5pt}

\vspace{-15pt}
\subsection{Trajectory Distribution}
To mechanistically explain these benchmark gains, we track the outcome distribution of rejection sampling across training stages (Table~\ref{tab:rejection_sampling_evolution}). From a global perspective (Full Set), the transition from cold-start model to Post-RL SFT model demonstrates a massive stabilization effect: the proportion of perfectly mastered queries (\textit{All Correct}) surges from 30.8\% to 55.6\%, while intractable queries (\textit{All Wrong}) are more than halved (12.7\% $\rightarrow$ 6.0\%). The Post-RL Self-Distillation stage further consolidates this stability, pushing \textit{All Correct} to 58.1\%. 
Crucially, in a targeted boundary analysis, we isolate the exact subset of ``All Wrong'' queries that the cold-start model completely failed to solve. Remarkably, the Post-RL SFT policy successfully generates at least one correct trajectory in 68.1\% of these previously unsolvable cases (58.9\% \textit{Mixed} + 9.2\% \textit{All Correct}). This provides definitive statistical proof that VRE does not merely memorize reasoning formats, but genuinely achieves a Boundary Breakthrough, enabling the Post-RL SFT stage to successfully salvage and distill these hard queires into reliable capabilities.
\vspace{-5pt}
\begin{table}[t]
\centering
\caption{\textbf{Distributions of Rollout Trajectories Across Training Stages.} \textit{Full Set} denotes the complete dataset used for RL optimization. \textit{Cold-Start ``All Wrong'' Only} refers to the specific subset of ``hard queries'' where the Cold-Start model failed across all sampled trajectories. This subset highlights the model's ability to breakthrough its initial cognitive boundaries.}
\label{tab:rejection_sampling_evolution}
\setlength{\tabcolsep}{8pt}
\resizebox{\linewidth}{!}{
\begin{tabular}{l c c c c}
\toprule
\textbf{Models} & \textbf{Evaluation Set} & \textbf{All Correct} ($\uparrow$) & \textbf{Mixed} & \textbf{All Wrong} ($\downarrow$) \\
\midrule
Cold-Start  & Full Set & 30.8\% & 56.5\% & 12.7\%  \\
RL Model & Full Set & 55.6\% & 38.4\% & 6.0\% \\
Post-RL SFT  & Full Set & 58.1\% & 36.2\% & 5.7\% \\
\midrule
Post-RL SFT  & \textit{Cold-Start ``All Wrong'' Only} & 9.2\% & 58.9\% & 31.9\% \\
\bottomrule
\end{tabular}
}
\vspace{-10pt}
\end{table}
\vspace{-5pt}
\subsection{Statistical Verification of the Reflection Mechanism.}
\label{sec:reflction_ana}
\vspace{-5pt}

Beyond macroscopic benchmark improvements, we employ information-theoretic metrics to rigorously validate two fundamental properties of our VRE framework: the training stability of same-family formatting, and the genuine utility of the generated visual reflections.

\paragraph{Distributional Affinity via KL Divergence.}

\begin{figure}[htbp]
    \centering
    \begin{subfigure}[b]{0.42\linewidth}
        \centering
        \includegraphics[width=\textwidth]{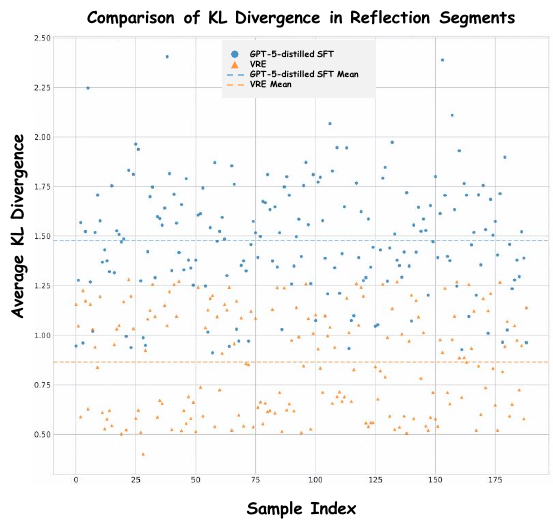}
        \caption{\textbf{Sample-level KL Divergence}}
        \label{fig:kl_a}
    \end{subfigure}
    \hfill
    \begin{subfigure}[b]{0.57\linewidth}
        \centering
        \includegraphics[width=\textwidth]{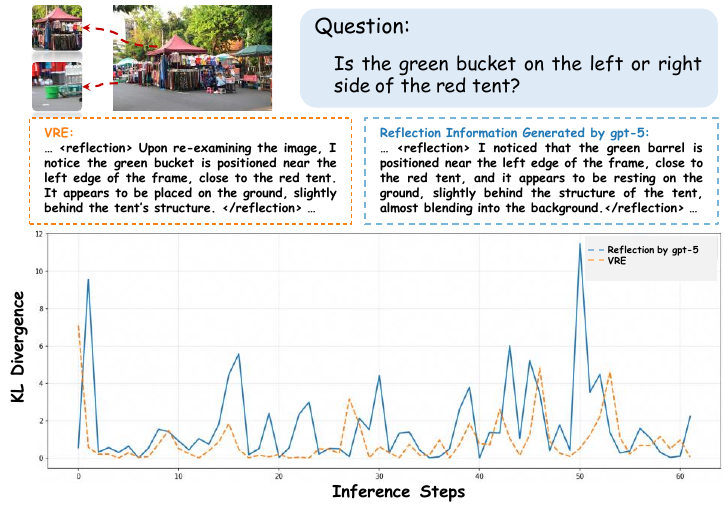}
        \caption{\textbf{Token-level KL Dynamics}}
        \label{fig:kl_b}
    \end{subfigure}
    
    \caption{\textbf{KL Divergence Analysis on $V^*$-Bench.} 
    \textbf{(a)} Sample-wise average KL divergence. 
    \textbf{(b)} A specific case study showing token-level KL divergence over inference steps. }
    \label{fig:kl_divergence}
    \vspace{-20pt}
\end{figure}
A critical design choice in VRE is utilizing the model's own policy distribution, rather than distilling traces from a stronger proprietary model (e.g., GPT-5), to construct the reflection dataset. To validate this necessity, we isolate the effect of data provenance by examining policies strictly after the cold-start SFT phase. 
Given a base MLLM $\pi_\theta$, we analyze the distribution shift by computing the Kullback-Leibler (KL) Divergence $D_{\text{KL}}( \pi_{\text{SFT}} \parallel \pi_\theta)$ on the $V^*$-Bench dataset, where $\pi_{\text{SFT}}$ represents obtained cold-start model. 
Specifically, we compare the divergence of a cold-start SFT model trained on GPT-5-distilled reflections ($\pi_{\text{SFT}}^{(gpt)}$) against an equivalent policy trained entirely on our self-generated VRE cold-start data ($\pi_{\text{SFT}}^{(0)}$).

As visualized in Fig.~\ref{fig:kl_divergence}(a), the $\pi_{\text{SFT}}^{(gpt)}$ exhibits a significantly higher average divergence across the entire dataset. Furthermore, the token-level case study (Fig.~\ref{fig:kl_divergence}(b)) reveals that it maintains consistently elevated KL values throughout the entire reflection generation process. This confirms that cross-model distillation forces the student to mimic alien reasoning patterns that deviate sharply from its native manifold. In contrast, the $\pi_{\text{SFT}}^{(0)}$ maintains a substantially lower KL divergence, ensuring that the acquired reflection capability is organically integrated and structurally compatible with the base model's intrinsic priors.
\vspace{-5pt}

\paragraph{Information Gain via Conditional Entropy.}

\begin{figure}[t]
\centering
\includegraphics[width=0.85\linewidth]{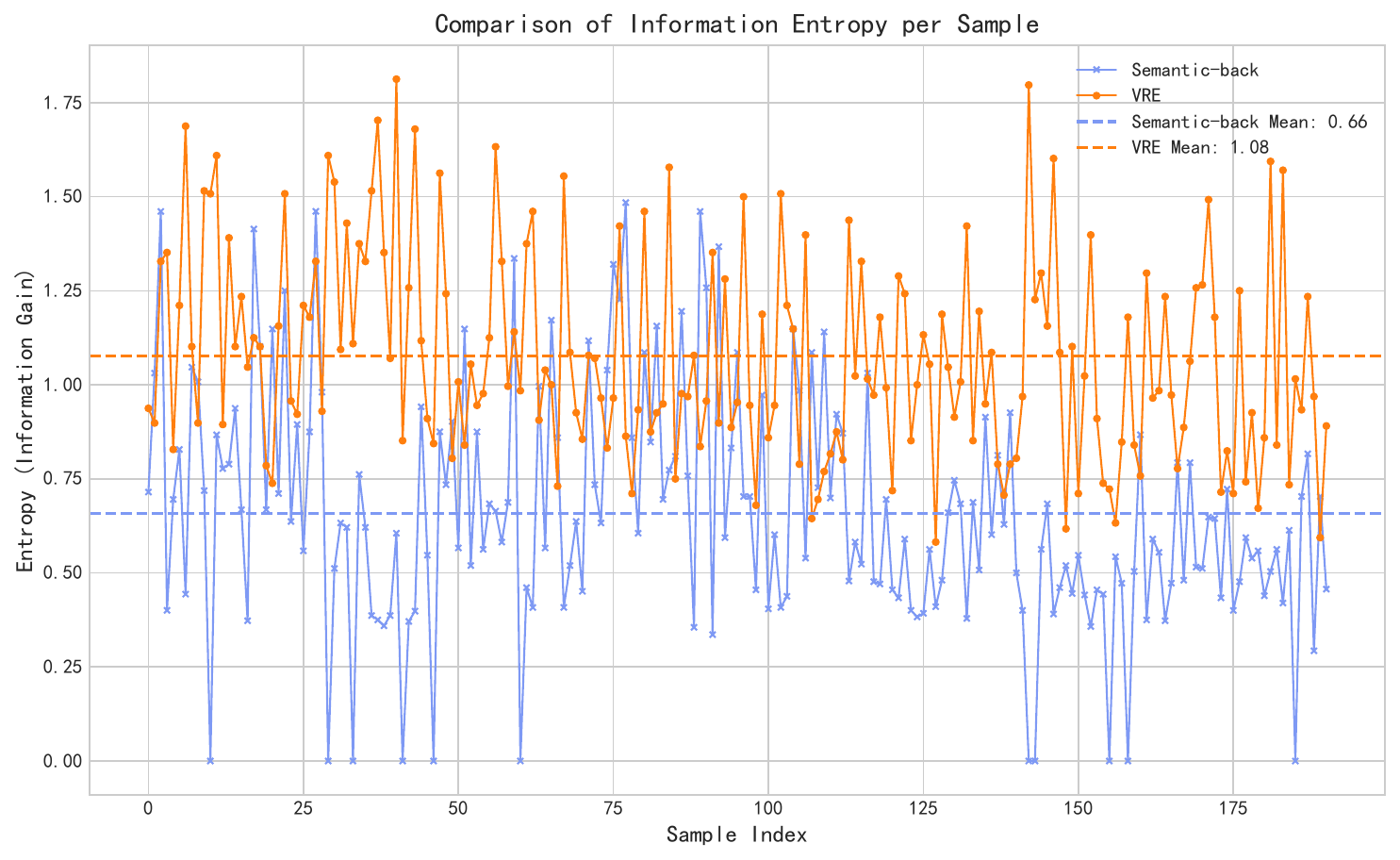}
\caption{\textbf{Information Entropy Comparison on $V^*$-Bench.} VRE consistently maintains higher conditional entropy than the Semantic-back model. The shift toward higher entropy values validates that VRE's reflection mechanism actively retrieves novel, non-redundant visual evidence.}
\label{fig:compare_max_entropy}
\vspace{-15pt}
\end{figure}
A rigorous validation is performed to confirm that our \texttt{<reflection>} block injects genuine visual evidence rather than merely paraphrasing the preceding context (redundancy), and we quantify Reflection Information Gain by employing Conditional Information Entropy as a proxy.

Formally, for a reflection trajectory $\tau = \{y_1, y_2, \dots, y_T\}$, we calculate its conditional entropy as the average Negative Log-Likelihood (NLL) of the model's output distribution conditioned on context $X_{\text{ctx}}$:
\begin{equation}
    H(\tau \mid X_{\text{ctx}}) = \frac{1}{T} \sum_{t=1}^{T} -\log \pi_\theta(y_t \mid X_{\text{ctx}}, y_{<t})
\end{equation}

where $\pi_\theta(y_t \mid X_{\text{ctx}}, y_{<t})$ denotes the conditional probability of token $y_t$ predicted by the policy $\pi_\theta$. In practice, this metric is computed using the CrossEntropyLoss between the model's logits and the ground-truth target tokens. 

A low entropy value indicates that the reflection is highly predictable given the textual context, suggesting the generation is redundant (``textual inertia''). Conversely, a high entropy score implies the introduction of external, non-redundant visual details—a proxy for the acquisition of novel information. As visualized in Figure~\ref{fig:compare_max_entropy}, VRE exhibits a significantly higher average entropy ($\bar{H}_{\text{VRE}} = 1.08$) compared to the Semantic-back~\cite{yang2025look} ($\bar{H}_{\text{Semantic-back}} = 0.66$). This substantial divergence confirms that VRE successfully breaks the textual inertia, compelling the model to retrieve targeted visual evidence rather than collapsing into degenerate self-repetition.

\vspace{-5pt}
\subsection{Training Dynamics and Stability}
\label{sec:training_dynamics}
\begin{figure}[htbp]
    \centering
    \vspace{-10pt}
    \begin{subfigure}[b]{0.48\linewidth}
        \centering
        \includegraphics[width=\textwidth]{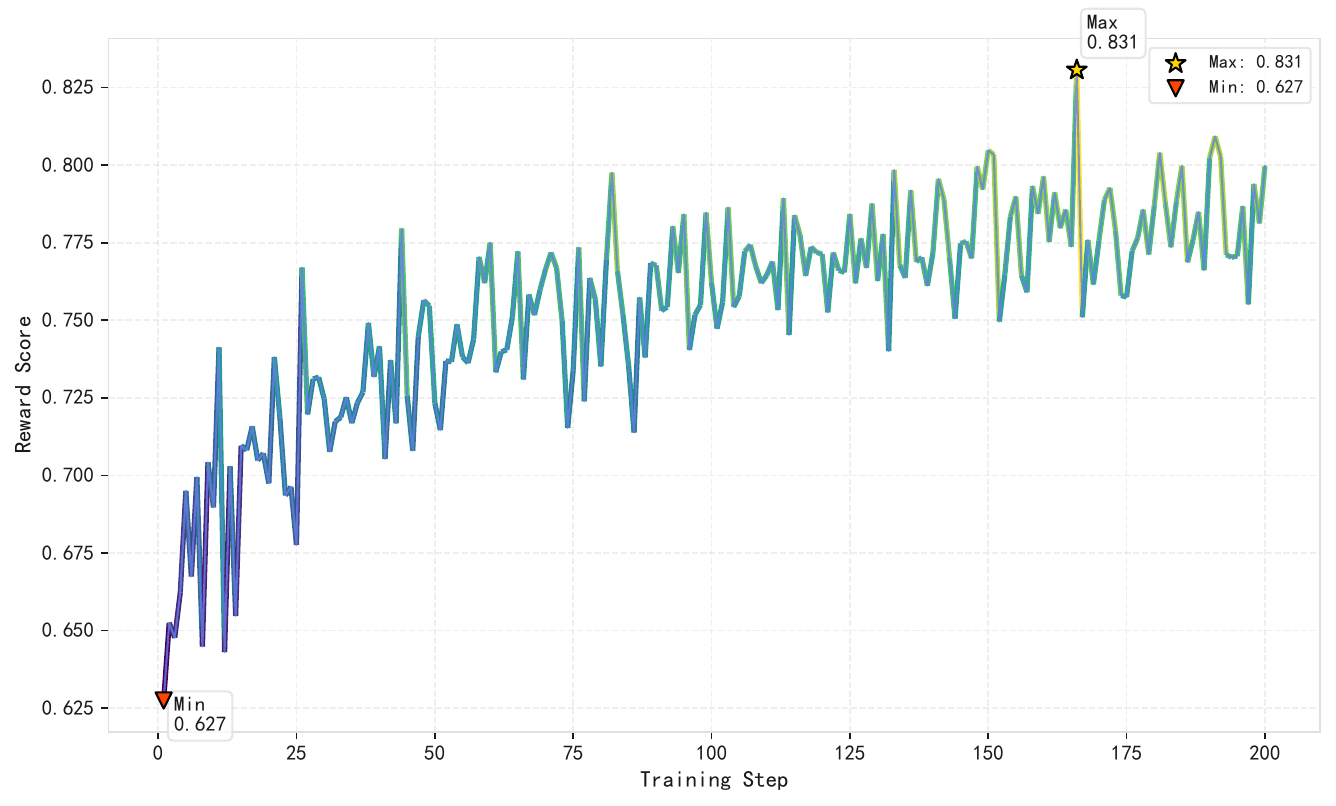}
        \caption{\textbf{Reward Score Evolution}}
        \label{fig:reward}
    \end{subfigure}
    \hfill
    \begin{subfigure}[b]{0.48\linewidth}
        \centering
        \includegraphics[width=\textwidth]{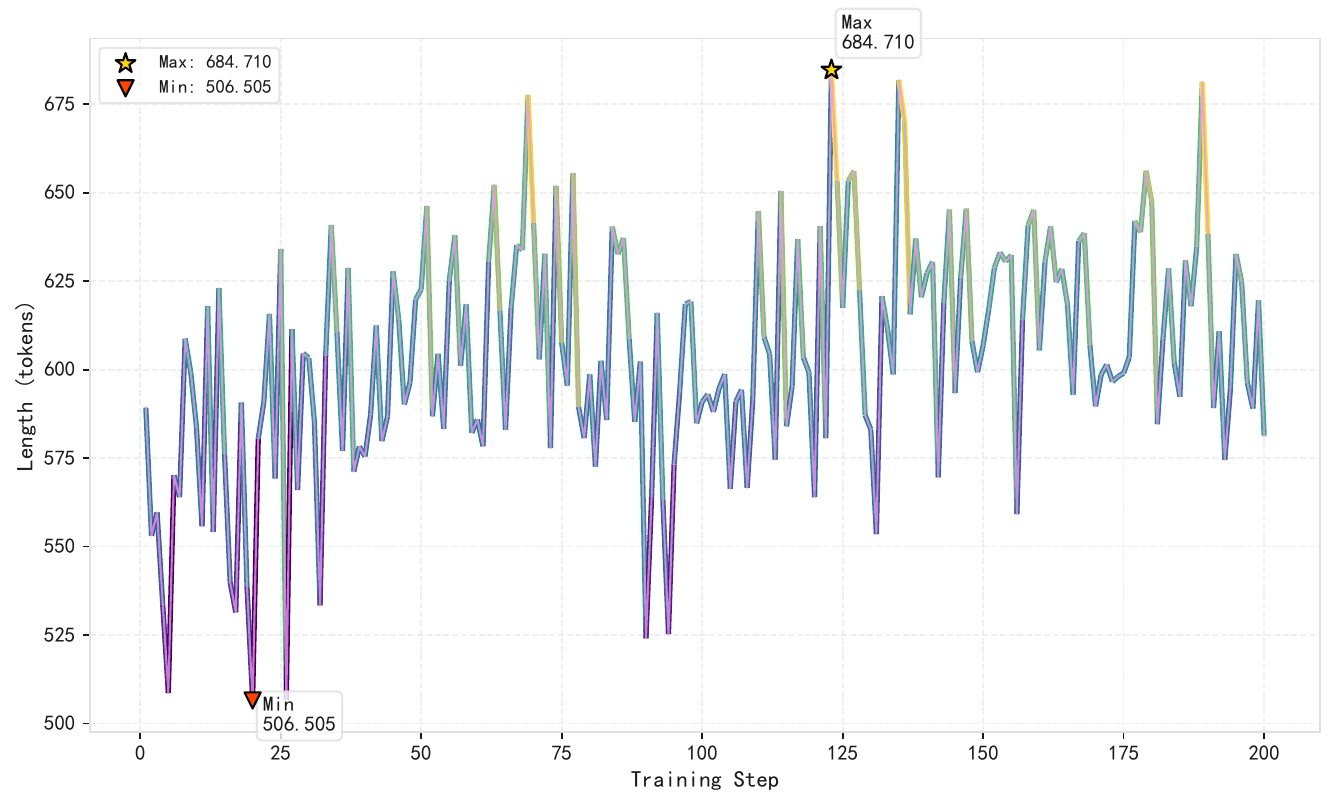}
        \caption{\textbf{Response Length Stability}}
        \label{fig:length}
    \end{subfigure}
    
    \caption{\textbf{Training Dynamics of GRPO.} 
    \textbf{(a)} The reward score steadily increases and converges around step 50, reflecting stable policy optimization. 
    \textbf{(b)} The average response length stabilizes at $\sim$600 tokens, indicating that the model extracts visual evidence efficiently without degenerating into verbose patterns.}
    \label{fig:rl_metrics}
\end{figure}

We visualize the trajectory of key optimization metrics during the GRPO phase in Fig.~\ref{fig:rl_metrics}. The curves reveal two critical insights regarding the stability and efficiency of our self-iterative strategy:

\begin{itemize}
    \item \textbf{Rapid Convergence via Reliable Self-Distillation:} As shown in Fig.~\ref{fig:rl_metrics}(a), the reward score exhibits a rapid ascent and converges effectively after approximately 50 steps. This accelerated alignment stems from the high distributional affinity of our self-evolution data. Unlike distilling from external proprietary models, our framework initializes the policy within its own native probability manifold. This allows the RL algorithm to focus immediately on optimizing reasoning paths rather than struggling to adapt to alien response distributions.
    
    \item \textbf{Mitigation of Reward Hacking (Information Density):} Crucially, the average response length (Fig.~\ref{fig:rl_metrics}(b)) stabilizes around 600 tokens rather than exhibiting unbounded growth. This stability mirrors the explicit constraints imposed during our data construction phase—which prioritized ``brief but information-rich'' reflections. It confirms that the model has internalized the objective of maximizing information density, successfully circumventing the common RL pitfall of gaming the reward system via verbose self-repetition or meaningless tautology.
\end{itemize}
\vspace{-15pt}

\vspace{-25pt}
\begin{figure}[h]
\centering
\includegraphics[width=0.5\linewidth]{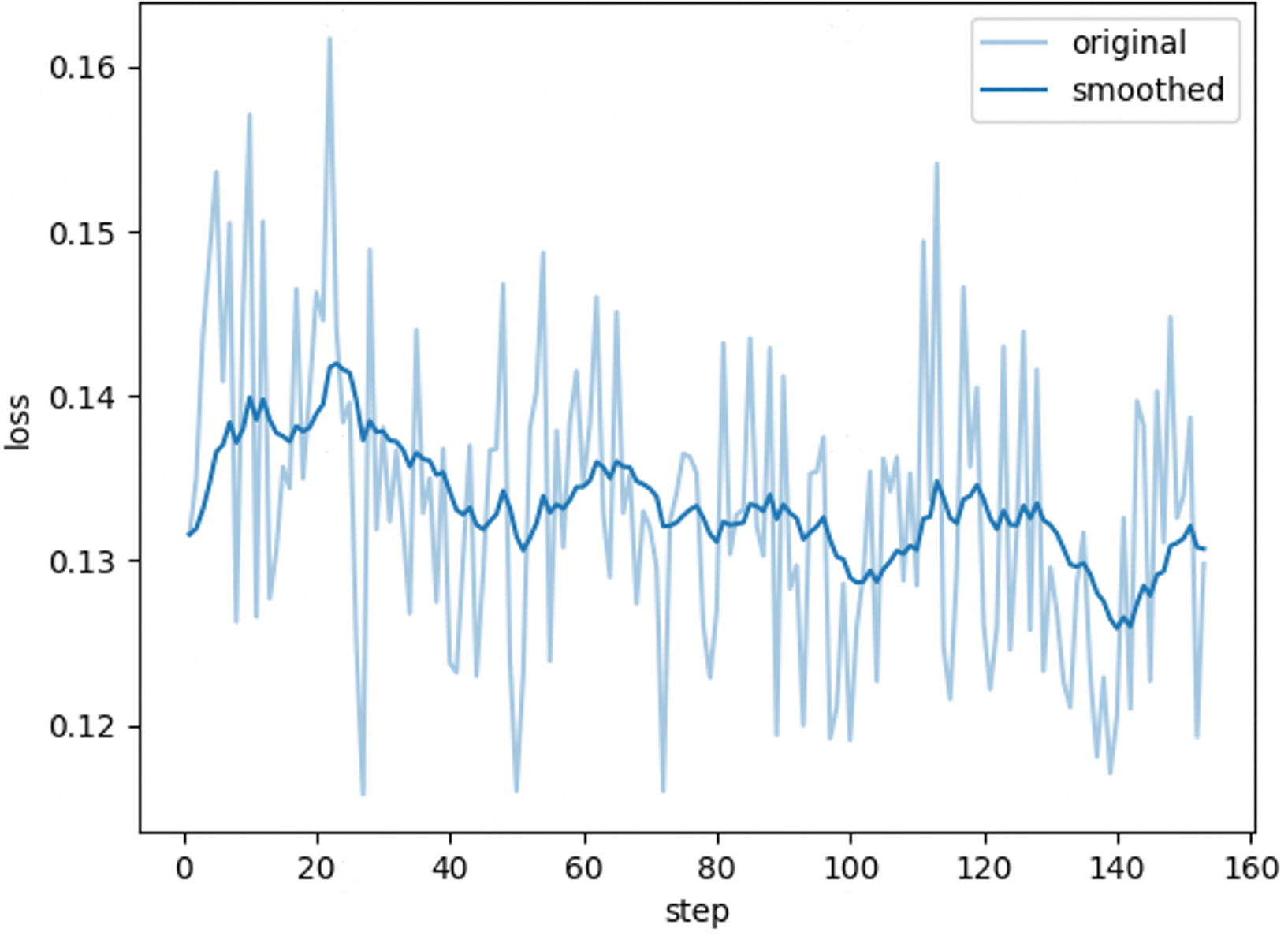}
\vspace{-5pt}
\caption{\textbf{Training Loss during Post-RL SFT.} 
}
\label{fig:iter_sft_loss}
\end{figure}
As shown in Fig.~\ref{fig:iter_sft_loss}, the loss curve during Post-RL SFT remains exceptionally stable with minimal fluctuation. This empirical observation yields a profound insight: because the training trajectories are self-generated by the RL-improved policy and rigorously filtered via rollout trajectories , the target data distribution aligns almost perfectly with the model's native probability manifold. Consequently, Post-RL SFT does not induce violent gradient shocks or catastrophic forgetting; rather, it smoothly and safely consolidates the exploratory gains on ``hard prompts'' into stable parametric memory, confirming the robustness of our closed-loop self-evolution paradigm.

\vspace{-10pt}
\subsection{Case}
We present four visualization samples of VRE across diverse scenarios. In particular, Sample 4 highlights the importance of post-RL SFT.

\begin{promptbox*}[Sample 1: VRE Generated Samples with Corrective Reflection
]{varcolor}{prompt:Sample_1}
\noindent
\begin{minipage}[c]{0.37\textwidth}
  \centering
  \vspace{-15pt}
  \includegraphics[width=\linewidth]{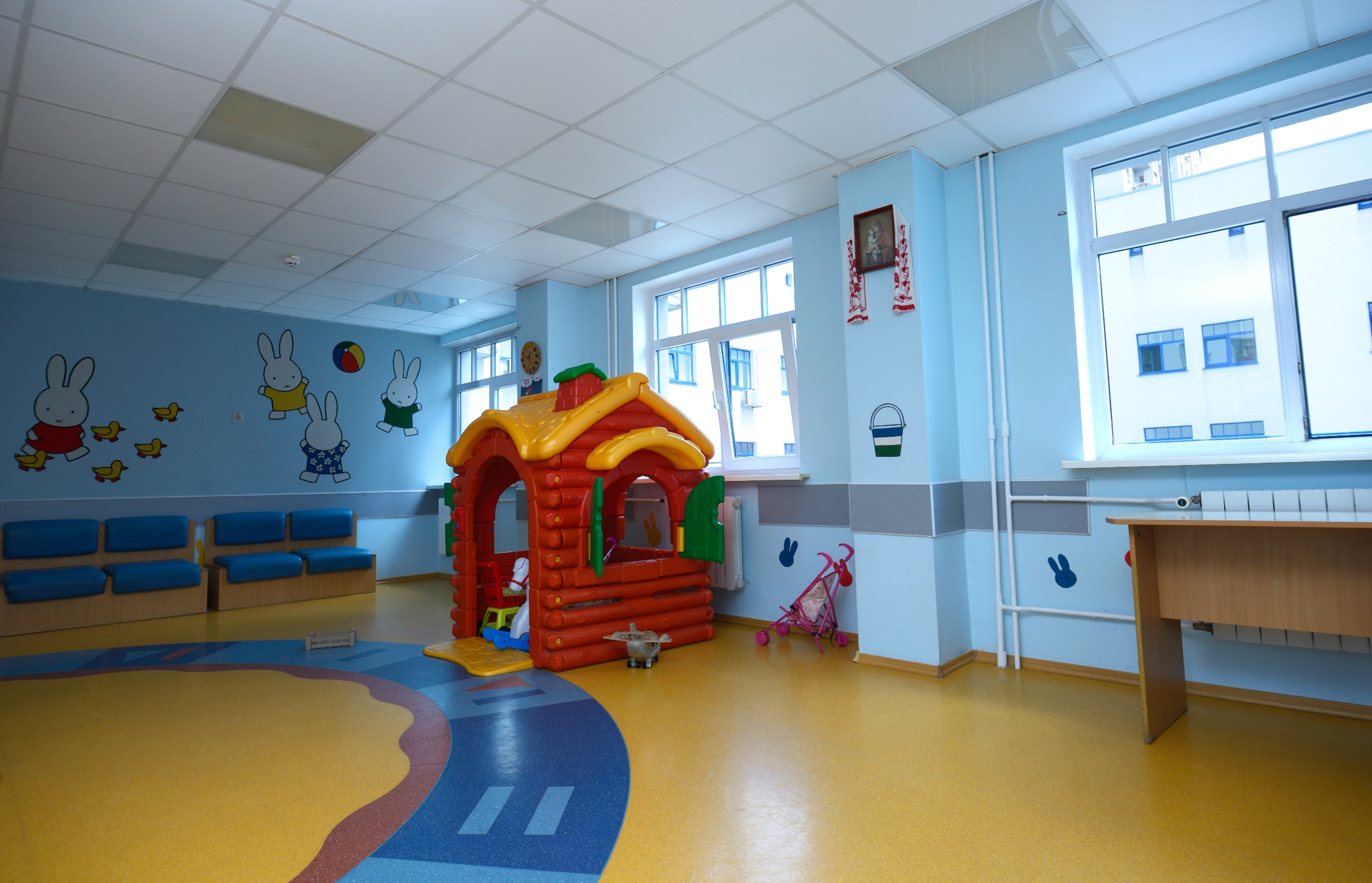}
\end{minipage}
\hfill
\begin{minipage}[c]{0.60\textwidth}
  \textbf{Question}: 
  How many people are in the oil painting?\\(A) 4 people\\(B) 0 people\\(C) 2 people\\(D) 1 people \\[4pt]
\noindent
\colorbox{gray!15}{%
  \parbox{\linewidth}{\raggedright\ttfamily
    Source: $V^*$-Bench\\\textbf{\textcolor{red}{Ground truth: C}}
    }%
}
\\

\end{minipage}

\vspace{6pt}
\begingroup\ttfamily


Let's analyze the image step by step:\\
1. The image is of a children's playroom.\\
2. There is a small oil painting hanging on the wall near the window.\\
3. In the painting, there appears to be one figure that looks like a person.\\ 
Given these observations, we can conclude that the oil painting contains 1 person. \textcolor{red}{(Incorrect.)}\\
\textbf{<reflection>} \\
Upon re-examining the image, I notice the oil painting more closely.\\ 
It seems to depict a figure holding a child, possibly a mother and baby.\\ 
This detail was not initially considered. \\
\textcolor{blue}{(Corrected using the image; correct answer obtained.)}\\
\textbf{</reflection>} \\[4pt]
The painting indeed shows two figures: an adult and a child.\\ 
Therefore, the number of people in the oil painting is 2.\\
\textbf{<answer>} C
\textbf{</answer>} 
\textcolor{blue}{(Correct.)}
\endgroup

\end{promptbox*}

\begin{promptbox*}[Sample 2: VRE Generated Samples with Recheck Reflection
]{varcolor}{prompt:Sample_2}
\noindent
\begin{minipage}[c]{0.37\textwidth}
  \centering
  \includegraphics[width=\linewidth]{figures/9204.jpg}
\end{minipage}
\hfill
\begin{minipage}[c]{0.60\textwidth}
  \textbf{Question}: 
  what is the number in the image? \\[4pt]
\noindent
\colorbox{gray!15}{%
  \parbox{\linewidth}{\raggedright\ttfamily
    Source: OCRBench\_V2\\\textbf{\textcolor{red}{Ground truth: 12988}}
    }%
}
\\

\end{minipage}

\vspace{6pt}
\begingroup\ttfamily

The image appears to show a handwritten number.\\ 
The digits seem to be "12988".\\
However, the first digit is partially obscured or unclear due to the handwriting style.
\textcolor{red}{(Not confident about the answer.)}\\
\textbf{<reflection>} \\
I will focus on checking whether the first digit is '1'.\\
Upon re-examining the image, the first digit is '1', then '2', and finally '988'.\\
The spacing between the digits is noticeable, which helps in distinguishing them clearly.\\ There are no other characters or numbers present.\\
\textcolor{blue}{(Corrected rechecking the image; correct answer obtained.)}\\
\textbf{</reflection>} \\[4pt]
Given the clear separation and the visible digits, the number in the image is indeed "12988". \\
\textbf{<answer>} 12988
\textbf{</answer>} 
\textcolor{blue}{(Correct.)}
\endgroup

\end{promptbox*}

\begin{promptbox*}[Sample 3: VRE Generated Samples with Elaborative Reflection
]{varcolor}{prompt:Sample_3}
\noindent
\begin{minipage}[c]{0.37\textwidth}
  \centering
  \includegraphics[width= \linewidth]{figures/two_col_1716.png}
\end{minipage}
\hfill
\begin{minipage}[c]{0.60\textwidth}
  \textbf{Question}: 
  Which year has the highest average annual wage? \\[4pt]
\noindent
\colorbox{gray!15}{%
  \parbox{\linewidth}{\raggedright\ttfamily
    Source: ChartQA\\\textbf{\textcolor{red}{Ground truth: 2019}}
    }%
}
\\

\end{minipage}

\vspace{6pt}
\begingroup\ttfamily
% ...

The graph shows the average annual wages in euros over several years.\\ To determine which year has the highest average annual wage, I need to look at the data points for each year and identify the highest value.\\ The data points are as follows:\\
- 2000: 31,247 euros
...
- 2019: 39,099 euros
\\
The highest value is 39,099 euros, which occurs in 2019.
\\
\textbf{<reflection>}
\\ Upon re-examining the image, I notice that the data point for 2019 is slightly higher than the one for 2018.\\
This detail was not initially considered in the previous reasoning.\\
\textcolor{blue}{(Elaborative reflection: enriched the reasoning with fine-grained spatial evidence.)}\\
\textbf{</reflection>} \\[4pt]
Given the updated observation, the highest average annual wage is indeed 39,099 euros, occurring in 2019.\\
The slight difference between 2018 and 2019 confirms that 2019 has the highest value.\\
\textbf{<answer>} 2019 
\textbf{</answer>} 
\textcolor{blue}{(Correct.)}
\endgroup

\end{promptbox*}

\vspace{-10pt}
\begin{promptbox*}[Sample 4: VRE Generated Samples with Corrective Reflection
]{varcolor}{prompt:Sample_4}
\noindent
\begin{minipage}[c]{0.37\textwidth}
  \centering
  \includegraphics[width=0.5\linewidth]{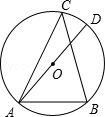}
\end{minipage}
\hfill
\begin{minipage}[c]{0.60\textwidth}
  \textbf{Question}: 
  As shown in the figure, the length of chord AB is 10.0, and the angle of circumference angle ACB = 45.0, then the diameter of the circle AD is () Choices:\\A: $5\sqrt{2}$   
B: $10\sqrt{2}$   
C: $15\sqrt{2}$   
D: $20\sqrt{2}$ \\[4pt]
\noindent
\colorbox{gray!15}{%
  \parbox{\linewidth}{\raggedright\ttfamily
    Source: MathVerse\\\textbf{\textcolor{red}{Ground truth: B}}
    }%
}
\\

\end{minipage}

\begingroup\ttfamily
\textcolor{Green}{\textbf{======================== RL Model: ========================}}\\
According to the given information, we can connect OA and OB. Since AB=10, we have OA=OB=5. Therefore, AD=$5\sqrt{2}$. Thus, the correct option is A.\textcolor{red}{(Incorrect.)}\\
\textbf{<reflection>} \\
Upon re-examining the image, I notice that the triangle AOB is formed by connecting points A, O, and B. The angle ACB is given as 45 degrees, which implies that angle AOB is twice that, making it 90 degrees. This confirms that triangle AOB is a right-angled triangle at O. \textcolor{red}{(Incomplete reflection: failed to detect logical conflict.)}\\
\textbf{</reflection>} \\[4pt]
Since AB = 10, we have OA = OB = 5. The diameter AD is twice the length of OA, so AD = 2 * 5 = 10. However, considering the right angle at O, the actual length of AD is $10\sqrt{2}$. Therefore, the correct answer is B. \\
\textbf{<answer>} B 
\textbf{</answer>} \textcolor{red}{(Forced the correct answer through broken and contradictory math logic.)}


\textcolor{Green}{\textbf{=================== Post-RL SFT  Model: ===================}}\\
According to the given information, we can connect OA and OB. Since angle ACB=45°, it follows that angle AOB=90°. Also, since AB=10, we have OA=OB=5. Therefore, AD=5$\sqrt{2}$. Thus, the correct option is A.\textcolor{red}{(Incorrect.)}\\
\textbf{<reflection>} \\
Upon re-examining the image, I notice that the triangle AOB is formed by connecting points A, O, and B. The angle ACB is given as 45 degrees, which implies that angle AOB is twice that, making it 90 degrees. This confirms that triangle AOB is a right-angled triangle at O. Additionally, the length of AB is provided as 10 units, if OA=OB=5, then Pythagorean theorem is not satisfied.\textcolor{blue}{(Corrective reflection: identified the exact mathematical contradiction.)}\\
\textbf{</reflection>} \\[4pt]
Given that triangle AOB is a right-angled triangle with AB as the hypotenuse, we can use the Pythagorean theorem to find the lengths of OA and OB. Since AB = 10, we have OA = OB = $5\sqrt{2}$. The diameter AD is twice the length of OA, so AD = 2 * $5\sqrt{2}$ = $10\sqrt{2}$. However, considering the right angle at O, the actual length of AD is $10\sqrt{2}$. Therefore, the correct answer is B.\\
\textbf{<answer>} B
\textbf{</answer>} 
\textcolor{blue}{(Correct.)}
\endgroup

\end{promptbox*}

\clearpage
\bibliographystyle{splncs04}
\bibliography{main}

@String(AAAI  = {AAAI})

@article{xu2024llava,
  title={Llava-o1: Let vision language models reason step-by-step},
  author={Xu, Guowei and Jin, Peng and Hao, Li and Song, Yibing and Sun, Lichao and Yuan, Li},
  journal={arXiv preprint arXiv:2411.10440},
  year={2024}
}

@article{liu2023visual,
  title={Visual instruction tuning},
  author={Liu, Haotian and Li, Chunyuan and Wu, Qingyang and Lee, Yong Jae},
  journal={Advances in neural information processing systems},
  volume={36},
  pages={34892--34916},
  year={2023}
}

@inproceedings{amizadeh2020neuro,
  title={Neuro-symbolic visual reasoning: Disentangling},
  author={Amizadeh, Saeed and Palangi, Hamid and Polozov, Alex and Huang, Yichen and Koishida, Kazuhito},
  booktitle={International Conference on Machine Learning},
  pages={279--290},
  year={2020},
  organization={Pmlr}
}

@article{garcez2019neural,
  title={Neural-symbolic computing: An effective methodology for principled integration of machine learning and reasoning},
  author={Garcez, Artur d'Avila and Gori, Marco and Lamb, Luis C and Serafini, Luciano and Spranger, Michael and Tran, Son N},
  journal={arXiv preprint arXiv:1905.06088},
  year={2019}
}

@inproceedings{gupta2023visual,
  title={Visual programming: Compositional visual reasoning without training},
  author={Gupta, Tanmay and Kembhavi, Aniruddha},
  booktitle={Proceedings of the IEEE/CVF Conference on Computer Vision and Pattern Recognition},
  pages={14953--14962},
  year={2023}
}

@article{lin2025uniworld,
  title={UniWorld: High-Resolution Semantic Encoders for Unified Visual Understanding and Generation},
  author={Lin, Bin and Li, Zongjian and Cheng, Xinhua and Niu, Yuwei and Ye, Yang and He, Xianyi and Yuan, Shenghai and Yu, Wangbo and Wang, Shaodong and Ge, Yunyang and others},
  journal={arXiv preprint arXiv:2506.03147},
  year={2025}
}

@article{liao2025langbridge,
  title={LangBridge: Interpreting Image as a Combination of Language Embeddings},
  author={Liao, Jiaqi and Niu, Yuwei and Meng, Fanqing and Li, Hao and Tian, Changyao and Du, Yinuo and Xiong, Yuwen and Li, Dianqi and Zhu, Xizhou and Yuan, Li and others},
  journal={arXiv preprint arXiv:2503.19404},
  year={2025}
}

@article{thawakar2025llamav,
  title={Llamav-o1: Rethinking step-by-step visual reasoning in llms},
  author={Thawakar, Omkar and Dissanayake, Dinura and More, Ketan and Thawkar, Ritesh and Heakl, Ahmed and Ahsan, Noor and Li, Yuhao and Zumri, Mohammed and Lahoud, Jean and Anwer, Rao Muhammad and others},
  journal={arXiv preprint arXiv:2501.06186},
  year={2025}
}

@article{guo2024mammoth,
  title={Mammoth-vl: Eliciting multimodal reasoning with instruction tuning at scale},
  author={Guo, Jarvis and Zheng, Tuney and Bai, Yuelin and Li, Bo and Wang, Yubo and Zhu, King and Li, Yizhi and Neubig, Graham and Chen, Wenhu and Yue, Xiang},
  journal={arXiv preprint arXiv:2412.05237},
  year={2024}
}

@article{bai2023qwen,
  title={Qwen-vl: A frontier large vision-language model with versatile abilities},
  author={Bai, Jinze and Bai, Shuai and Yang, Shusheng and Wang, Shijie and Tan, Sinan and Wang, Peng and Lin, Junyang and Zhou, Chang and Zhou, Jingren},
  journal={arXiv preprint arXiv:2308.12966},
  volume={1},
  number={2},
  pages={3},
  year={2023}
}

@article{hurst2024gpt,
  title={Gpt-4o system card},
  author={Hurst, Aaron and Lerer, Adam and Goucher, Adam P and Perelman, Adam and Ramesh, Aditya and Clark, Aidan and Ostrow, AJ and Welihinda, Akila and Hayes, Alan and Radford, Alec and others},
  journal={arXiv preprint arXiv:2410.21276},
  year={2024}
}

@article{shao2024deepseekmath,
  title={Deepseekmath: Pushing the limits of mathematical reasoning in open language models},
  author={Shao, Zhihong and Wang, Peiyi and Zhu, Qihao and Xu, Runxin and Song, Junxiao and Bi, Xiao and Zhang, Haowei and Zhang, Mingchuan and Li, YK and Wu, Y and others},
  journal={arXiv preprint arXiv:2402.03300},
  year={2024}
}

@misc{meng2025mmeureka,
    title={MM-Eureka: Exploring the Frontiers of Multimodal Reasoning with Rule-based Reinforcement Learning},
    author={Fanqing Meng and Lingxiao Du and Zongkai Liu and Zhixiang Zhou and Quanfeng Lu and Daocheng Fu and Tiancheng Han and Botian Shi and Wenhai Wang and Junjun He and Kaipeng Zhang and Ping Luo and Yu Qiao and Qiaosheng Zhang and Wenqi Shao},
    year={2025},
    eprint={2503.07365},
    archivePrefix={arXiv},
    primaryClass={cs.CV}
}

@article{jaech2024openai,
  title={Openai o1 system card},
  author={Jaech, Aaron and Kalai, Adam and Lerer, Adam and Richardson, Adam and El-Kishky, Ahmed and Low, Aiden and Helyar, Alec and Madry, Aleksander and Beutel, Alex and Carney, Alex and others},
  journal={arXiv preprint arXiv:2412.16720},
  year={2024}
}

@article{liu2025visual,
  title={Visual-rft: Visual reinforcement fine-tuning},
  author={Liu, Ziyu and Sun, Zeyi and Zang, Yuhang and Dong, Xiaoyi and Cao, Yuhang and Duan, Haodong and Lin, Dahua and Wang, Jiaqi},
  journal={arXiv preprint arXiv:2503.01785},
  year={2025}
}

@article{zhang2025chain,
  title={Chain-of-Focus: Adaptive Visual Search and Zooming for Multimodal Reasoning via RL},
  author={Zhang, Xintong and Gao, Zhi and Zhang, Bofei and Li, Pengxiang and Zhang, Xiaowen and Liu, Yang and Yuan, Tao and Wu, Yuwei and Jia, Yunde and Zhu, Song-Chun and others},
  journal={arXiv preprint arXiv:2505.15436},
  year={2025}
}

@article{su2025openthinkimg,
  title={Openthinkimg: Learning to think with images via visual tool reinforcement learning},
  author={Su, Zhaochen and Li, Linjie and Song, Mingyang and Hao, Yunzhuo and Yang, Zhengyuan and Zhang, Jun and Chen, Guanjie and Gu, Jiawei and Li, Juntao and Qu, Xiaoye and others},
  journal={arXiv preprint arXiv:2505.08617},
  year={2025}
}

@article{chern2025thinking,
  title={Thinking with Generated Images},
  author={Chern, Ethan and Hu, Zhulin and Chern, Steffi and Kou, Siqi and Su, Jiadi and Ma, Yan and Deng, Zhijie and Liu, Pengfei},
  journal={arXiv preprint arXiv:2505.22525},
  year={2025}
}

@misc{qwen2.5-VL,
    title = {Qwen2.5-VL},
    author = {Qwen Team},
    month = {January},
    year = {2025}
}

@article{Qwen2VL,
  title={Qwen2-VL: Enhancing Vision-Language Model's Perception of the World at Any Resolution},
  author={Wang, Peng and Bai, Shuai and Tan, Sinan and Wang, Shijie and Fan, Zhihao and Bai, Jinze and Chen, Keqin and Liu, Xuejing and Wang, Jialin and Ge, Wenbin and Fan, Yang and Dang, Kai and Du, Mengfei and Ren, Xuancheng and Men, Rui and Liu, Dayiheng and Zhou, Chang and Zhou, Jingren and Lin, Junyang},
  journal={arXiv preprint arXiv:2409.12191},
  year={2024}
}

@inproceedings{zhang2024mathverse,
  title={Mathverse: Does your multi-modal llm truly see the diagrams in visual math problems?},
  author={Zhang, Renrui and Jiang, Dongzhi and Zhang, Yichi and Lin, Haokun and Guo, Ziyu and Qiu, Pengshuo and Zhou, Aojun and Lu, Pan and Chang, Kai-Wei and Qiao, Yu and others},
  booktitle={European Conference on Computer Vision},
  pages={169--186},
  year={2024},
  organization={Springer}
}

@misc{qwen2025qwen25technicalreport,
      title={Qwen2.5 Technical Report}, 
      author={Qwen and : and An Yang and Baosong Yang and Beichen Zhang and Binyuan Hui and Bo Zheng and Bowen Yu and Chengyuan Li and Dayiheng Liu and Fei Huang and Haoran Wei and Huan Lin and Jian Yang and Jianhong Tu and Jianwei Zhang and Jianxin Yang and Jiaxi Yang and Jingren Zhou and Junyang Lin and Kai Dang and Keming Lu and Keqin Bao and Kexin Yang and Le Yu and Mei Li and Mingfeng Xue and Pei Zhang and Qin Zhu and Rui Men and Runji Lin and Tianhao Li and Tianyi Tang and Tingyu Xia and Xingzhang Ren and Xuancheng Ren and Yang Fan and Yang Su and Yichang Zhang and Yu Wan and Yuqiong Liu and Zeyu Cui and Zhenru Zhang and Zihan Qiu},
      year={2025},
      eprint={2412.15115},
      archivePrefix={arXiv},
      primaryClass={cs.CL},
}

@article{wang2024measuring,
  title={Measuring multimodal mathematical reasoning with math-vision dataset},
  author={Wang, Ke and Pan, Junting and Shi, Weikang and Lu, Zimu and Ren, Houxing and Zhou, Aojun and Zhan, Mingjie and Li, Hongsheng},
  journal={Advances in Neural Information Processing Systems},
  volume={37},
  pages={95095--95169},
  year={2024}
}

@article{qiao2024we,
  title={We-math: Does your large multimodal model achieve human-like mathematical reasoning?},
  author={Qiao, Runqi and Tan, Qiuna and Dong, Guanting and Wu, Minhui and Sun, Chong and Song, Xiaoshuai and GongQue, Zhuoma and Lei, Shanglin and Wei, Zhe and Zhang, Miaoxuan and others},
  journal={arXiv preprint arXiv:2407.01284},
  year={2024}
}

@inproceedings{guan2024hallusionbench,
  title={Hallusionbench: an advanced diagnostic suite for entangled language hallucination and visual illusion in large vision-language models},
  author={Guan, Tianrui and Liu, Fuxiao and Wu, Xiyang and Xian, Ruiqi and Li, Zongxia and Liu, Xiaoyu and Wang, Xijun and Chen, Lichang and Huang, Furong and Yacoob, Yaser and others},
  booktitle={Proceedings of the IEEE/CVF Conference on Computer Vision and Pattern Recognition},
  pages={14375--14385},
  year={2024}
}

@article{zhu2025internvl3,
  title={Internvl3: Exploring advanced training and test-time recipes for open-source multimodal models},
  author={Zhu, Jinguo and Wang, Weiyun and Chen, Zhe and Liu, Zhaoyang and Ye, Shenglong and Gu, Lixin and Tian, Hao and Duan, Yuchen and Su, Weijie and Shao, Jie and others},
  journal={arXiv preprint arXiv:2504.10479},
  year={2025}
}

@article{zhang2025r1,
  title={R1-vl: Learning to reason with multimodal large language models via step-wise group relative policy optimization},
  author={Zhang, Jingyi and Huang, Jiaxing and Yao, Huanjin and Liu, Shunyu and Zhang, Xikun and Lu, Shijian and Tao, Dacheng},
  journal={arXiv preprint arXiv:2503.12937},
  year={2025}
}

@article{wang2025vl,
  title={Vl-rethinker: Incentivizing self-reflection of vision-language models with reinforcement learning},
  author={Wang, Haozhe and Qu, Chao and Huang, Zuming and Chu, Wei and Lin, Fangzhen and Chen, Wenhu},
  journal={arXiv preprint arXiv:2504.08837},
  year={2025}
}

@article{yang2025r1,
  title={R1-onevision: Advancing generalized multimodal reasoning through cross-modal formalization},
  author={Yang, Yi and He, Xiaoxuan and Pan, Hongkun and Jiang, Xiyan and Deng, Yan and Yang, Xingtao and Lu, Haoyu and Yin, Dacheng and Rao, Fengyun and Zhu, Minfeng and others},
  journal={arXiv preprint arXiv:2503.10615},
  year={2025}
}

@article{zhang2023multimodal,
  title={Multimodal chain-of-thought reasoning in language models},
  author={Zhang, Zhuosheng and Zhang, Aston and Li, Mu and Zhao, Hai and Karypis, George and Smola, Alex},
  journal={arXiv preprint arXiv:2302.00923},
  year={2023}
}

@article{hu2024visual,
  title={Visual sketchpad: Sketching as a visual chain of thought for multimodal language models},
  author={Hu, Yushi and Shi, Weijia and Fu, Xingyu and Roth, Dan and Ostendorf, Mari and Zettlemoyer, Luke and Smith, Noah A and Krishna, Ranjay},
  journal={arXiv preprint arXiv:2406.09403},
  year={2024}
}

@inproceedings{wu2024v,
  title={V*: Guided visual search as a core mechanism in multimodal llms},
  author={Wu, Penghao and Xie, Saining},
  booktitle={Proceedings of the IEEE/CVF Conference on Computer Vision and Pattern Recognition},
  pages={13084--13094},
  year={2024}
}

@inproceedings{li2025dyfo,
  title={Dyfo: A training-free dynamic focus visual search for enhancing lmms in fine-grained visual understanding},
  author={Li, Geng and Xu, Jinglin and Zhao, Yunzhen and Peng, Yuxin},
  booktitle={Proceedings of the Computer Vision and Pattern Recognition Conference},
  pages={9098--9108},
  year={2025}
}

@article{yang2025look,
  title={Look-back: Implicit visual re-focusing in mllm reasoning},
  author={Yang, Shuo and Niu, Yuwei and Liu, Yuyang and Ye, Yang and Lin, Bin and Yuan, Li},
  journal={arXiv preprint arXiv:2507.03019},
  year={2025}
}

@article{wang2025visuothink,
  title={Visuothink: Empowering lvlm reasoning with multimodal tree search},
  author={Wang, Yikun and Wang, Siyin and Cheng, Qinyuan and Fei, Zhaoye and Ding, Liang and Guo, Qipeng and Tao, Dacheng and Qiu, Xipeng},
  journal={arXiv preprint arXiv:2504.09130},
  year={2025}
}

@article{zhang2025thyme,
  title={Thyme: Think beyond images},
  author={Zhang, Yi-Fan and Lu, Xingyu and Yin, Shukang and Fu, Chaoyou and Chen, Wei and Hu, Xiao and Wen, Bin and Jiang, Kaiyu and Liu, Changyi and Zhang, Tianke and others},
  journal={arXiv preprint arXiv:2508.11630},
  year={2025}
}

@inproceedings{tong2024eyes,
  title={Eyes wide shut? exploring the visual shortcomings of multimodal llms},
  author={Tong, Shengbang and Liu, Zhuang and Zhai, Yuexiang and Ma, Yi and LeCun, Yann and Xie, Saining},
  booktitle={Proceedings of the IEEE/CVF conference on computer vision and pattern recognition},
  pages={9568--9578},
  year={2024}
}

@article{wu2024accelerating,
  title={Accelerating multimodal large language models via dynamic visual-token exit and the empirical findings},
  author={Wu, Qiong and Lin, Wenhao and Zhou, Yiyi and Ye, Weihao and Zen, Zhanpeng and Sun, Xiaoshuai and Ji, Rongrong},
  journal={arXiv preprint arXiv:2411.19628},
  year={2024}
}

@inproceedings{lin2025boosting,
  title={Boosting multimodal large language models with visual tokens withdrawal for rapid inference},
  author={Lin, Zhihang and Lin, Mingbao and Lin, Luxi and Ji, Rongrong},
  booktitle={Proceedings of the AAAI Conference on Artificial Intelligence},
  volume={39},
  pages={5334--5342},
  year={2025}
}

@article{wu2024controlmllm,
  title={Controlmllm: Training-free visual prompt learning for multimodal large language models},
  author={Wu, Mingrui and Cai, Xinyue and Ji, Jiayi and Li, Jiale and Huang, Oucheng and Luo, Gen and Fei, Hao and Jiang, Guannan and Sun, Xiaoshuai and Ji, Rongrong},
  journal={Advances in Neural Information Processing Systems},
  volume={37},
  pages={45206--45234},
  year={2024}
}

@article{deng2025openvlthinker,
  title={Openvlthinker: Complex vision-language reasoning via iterative sft-rl cycles},
  author={Deng, Yihe and Bansal, Hritik and Yin, Fan and Peng, Nanyun and Wang, Wei and Chang, Kai-Wei},
  journal={arXiv preprint arXiv:2503.17352},
  year={2025}
}

@misc{yeo2025demystifying,
      title={Demystifying Long Chain-of-Thought Reasoning in LLMs}, 
      author={Edward Yeo and Yuxuan Tong and Morry Niu and Graham Neubig and Xiang Yue},
      year={2025},
      eprint={2502.03373},
      archivePrefix={arXiv},
      primaryClass={cs.CL},
}

@article{hong2025deepeyesv2,
  title={Deepeyesv2: Toward agentic multimodal model},
  author={Hong, Jack and Zhao, Chenxiao and Zhu, ChengLin and Lu, Weiheng and Xu, Guohai and Yu, Xing},
  journal={arXiv preprint arXiv:2511.05271},
  year={2025}
}

@article{zheng2025deepeyes,
  title={Deepeyes: Incentivizing" thinking with images" via reinforcement learning},
  author={Zheng, Ziwei and Yang, Michael and Hong, Jack and Zhao, Chenxiao and Xu, Guohai and Yang, Le and Shen, Chao and Yu, Xing},
  journal={arXiv preprint arXiv:2505.14362},
  year={2025}
}

@article{liu2025more,
  title={More thinking, less seeing? assessing amplified hallucination in multimodal reasoning models},
  author={Liu, Chengzhi and Xu, Zhongxing and Wei, Qingyue and Wu, Juncheng and Zou, James and Wang, Xin Eric and Zhou, Yuyin and Liu, Sheng},
  journal={arXiv preprint arXiv:2505.21523},
  year={2025}
}

@article{li2024llava,
  title={Llava-onevision: Easy visual task transfer},
  author={Li, Bo and Zhang, Yuanhan and Guo, Dong and Zhang, Renrui and Li, Feng and Zhang, Hao and Zhang, Kaichen and Zhang, Peiyuan and Li, Yanwei and Liu, Ziwei and others},
  journal={arXiv preprint arXiv:2408.03326},
  year={2024}
}

@inproceedings{gao2023scaling,
  title={Scaling laws for reward model overoptimization},
  author={Gao, Leo and Schulman, John and Hilton, Jacob},
  booktitle={International Conference on Machine Learning},
  pages={10835--10866},
  year={2023},
  organization={PMLR}
}

@article{casper2023open,
  title={Open problems and fundamental limitations of reinforcement learning from human feedback},
  author={Casper, Stephen and Davies, Xander and Shi, Claudia and Gilbert, Thomas Krendl and Scheurer, J{\'e}r{\'e}my and Rando, Javier and Freedman, Rachel and Korbak, Tomasz and Lindner, David and Freire, Pedro and others},
  journal={arXiv preprint arXiv:2307.15217},
  year={2023}
}

@article{zhai2023investigating,
  title={Investigating the catastrophic forgetting in multimodal large language models},
  author={Zhai, Yuexiang and Tong, Shengbang and Li, Xiao and Cai, Mu and Qu, Qing and Lee, Yong Jae and Ma, Yi},
  journal={arXiv preprint arXiv:2309.10313},
  year={2023}
}

@article{rlhfanthropic,
  title={Training a helpful and harmless assistant with reinforcement learning from human feedback},
  author={Bai, Yuntao and Jones, Andy and Ndousse, Kamal and Askell, Amanda and Chen, Anna and DasSarma, Nova and Drain, Dawn and Fort, Stanislav and Ganguli, Deep and Henighan, Tom and others},
  journal={arXiv preprint arXiv:2204.05862},
  year={2022}
}

@article{oairlhf,
  title={Training language models to follow instructions with human feedback},
  author={Ouyang, Long and Wu, Jeffrey and Jiang, Xu and Almeida, Diogo and Wainwright, Carroll and Mishkin, Pamela and Zhang, Chong and Agarwal, Sandhini and Slama, Katarina and Ray, Alex and others},
  journal={Advances in neural information processing systems},
  volume={35},
  pages={27730--27744},
  year={2022}
}

@article{chen2024expanding,
  title={Expanding performance boundaries of open-source multimodal models with model, data, and test-time scaling},
  author={Chen, Zhe and Wang, Weiyun and Cao, Yue and Liu, Yangzhou and Gao, Zhangwei and Cui, Erfei and Zhu, Jinguo and Ye, Shenglong and Tian, Hao and Liu, Zhaoyang and others},
  journal={arXiv preprint arXiv:2412.05271},
  year={2024}
}

@article{achiam2023gpt,
  title={Gpt-4 technical report},
  author={Achiam, Josh and Adler, Steven and Agarwal, Sandhini and Ahmad, Lama and Akkaya, Ilge and Aleman, Florencia Leoni and Almeida, Diogo and Altenschmidt, Janko and Altman, Sam and Anadkat, Shyamal and others},
  journal={arXiv preprint arXiv:2303.08774},
  year={2023}
}

@article{chen2025mint,
  title={Mint-cot: Enabling interleaved visual tokens in mathematical chain-of-thought reasoning},
  author={Chen, Xinyan and Zhang, Renrui and Jiang, Dongzhi and Zhou, Aojun and Yan, Shilin and Lin, Weifeng and Li, Hongsheng},
  journal={arXiv preprint arXiv:2506.05331},
  year={2025}
}

@inproceedings{cai2024vip,
  title={Vip-llava: Making large multimodal models understand arbitrary visual prompts},
  author={Cai, Mu and Liu, Haotian and Mustikovela, Siva Karthik and Meyer, Gregory P and Chai, Yuning and Park, Dennis and Lee, Yong Jae},
  booktitle={Proceedings of the IEEE/CVF Conference on Computer Vision and Pattern Recognition},
  pages={12914--12923},
  year={2024}
}

@article{yang2023set,
  title={Set-of-mark prompting unleashes extraordinary visual grounding in gpt-4v},
  author={Yang, Jianwei and Zhang, Hao and Li, Feng and Zou, Xueyan and Li, Chunyuan and Gao, Jianfeng},
  journal={arXiv preprint arXiv:2310.11441},
  year={2023}
}

@article{meng2025mm,
  title={MM-Eureka: Exploring the Frontiers of Multimodal Reasoning with Rule-based Reinforcement Learning},
  author={Meng, Fanqing and Du, Lingxiao and Liu, Zongkai and Zhou, Zhixiang and Lu, Quanfeng and Fu, Daocheng and Han, Tiancheng and Shi, Botian and Wang, Wenhai and He, Junjun and others},
  journal={arXiv preprint arXiv:2503.07365},
  year={2025}
}

@article{tulu3,
  title={Tulu 3: Pushing frontiers in open language model post-training},
  author={Lambert, Nathan and Morrison, Jacob and Pyatkin, Valentina and Huang, Shengyi and Ivison, Hamish and Brahman, Faeze and Miranda, Lester James V and Liu, Alisa and Dziri, Nouha and Lyu, Shane and others},
  journal={arXiv preprint arXiv:2411.15124},
  year={2024}
}

@article{GRPO,
  title={Deepseekmath: Pushing the limits of mathematical reasoning in open language models},
  author={Shao, Zhihong and Wang, Peiyi and Zhu, Qihao and Xu, Runxin and Song, Junxiao and Bi, Xiao and Zhang, Haowei and Zhang, Mingchuan and Li, YK and Wu, Y and others},
  journal={arXiv preprint arXiv:2402.03300},
  year={2024}
}

@article{guo2025deepseek,
  title={Deepseek-r1: Incentivizing reasoning capability in llms via reinforcement learning},
  author={Guo, Daya and Yang, Dejian and Zhang, Haowei and Song, Junxiao and Zhang, Ruoyu and Xu, Runxin and Zhu, Qihao and Ma, Shirong and Wang, Peiyi and Bi, Xiao and others},
  journal={arXiv preprint arXiv:2501.12948},
  year={2025}
}

@article{bai2025qwen2,
  title={Qwen2. 5-vl technical report},
  author={Bai, Shuai and Chen, Keqin and Liu, Xuejing and Wang, Jialin and Ge, Wenbin and Song, Sibo and Dang, Kai and Wang, Peng and Wang, Shijie and Tang, Jun and others},
  journal={arXiv preprint arXiv:2502.13923},
  year={2025}
}

@misc{claude3.7sonnet,
  author       = {Anthropic},
  title        = {Claude 3.7 Sonnet System Card},
  year         = {2025}
}

@article{lu2023mathvista,
  title={Mathvista: Evaluating mathematical reasoning of foundation models in visual contexts},
  author={Lu, Pan and Bansal, Hritik and Xia, Tony and Liu, Jiacheng and Li, Chunyuan and Hajishirzi, Hannaneh and Cheng, Hao and Chang, Kai-Wei and Galley, Michel and Gao, Jianfeng},
  journal={arXiv preprint arXiv:2310.02255},
  year={2023}
}

@inproceedings{yue2024mmmu,
  title={Mmmu: A massive multi-discipline multimodal understanding and reasoning benchmark for expert agi},
  author={Yue, Xiang and Ni, Yuansheng and Zhang, Kai and Zheng, Tianyu and Liu, Ruoqi and Zhang, Ge and Stevens, Samuel and Jiang, Dongfu and Ren, Weiming and Sun, Yuxuan and others},
  booktitle={Proceedings of the IEEE/CVF Conference on Computer Vision and Pattern Recognition},
  pages={9556--9567},
  year={2024}
}

@article{fu2024ocrbench,
  title={Ocrbench v2: An improved benchmark for evaluating large multimodal models on visual text localization and reasoning},
  author={Fu, Ling and Kuang, Zhebin and Song, Jiajun and Huang, Mingxin and Yang, Biao and Li, Yuzhe and Zhu, Linghao and Luo, Qidi and Wang, Xinyu and Lu, Hao and others},
  journal={arXiv preprint arXiv:2501.00321},
  year={2024}
}

@inproceedings{masry2022chartqa,
  title={Chartqa: A benchmark for question answering about charts with visual and logical reasoning},
  author={Masry, Ahmed and Do, Xuan Long and Tan, Jia Qing and Joty, Shafiq and Hoque, Enamul},
  booktitle={Findings of the association for computational linguistics: ACL 2022},
  pages={2263--2279},
  year={2022}
}

@article{xiao2024logicvista,
  title={Logicvista: Multimodal llm logical reasoning benchmark in visual contexts},
  author={Xiao, Yijia and Sun, Edward and Liu, Tianyu and Wang, Wei},
  journal={arXiv preprint arXiv:2407.04973},
  year={2024}
}

\end{document}